\documentclass[preprint,12pt,authoryear]{elsarticle}

\usepackage{amssymb}
\usepackage{amsmath}

\usepackage[ruled,linesnumbered]{algorithm2e}
\usepackage{amsmath,amsfonts}
\usepackage{xcolor}
\usepackage{soul}
\usepackage{array}
\usepackage[caption=false,font=normalsize,labelfont=rm,textfont=rm]{subfig}
\usepackage{textcomp}
\usepackage{stfloats}
\usepackage{url}
\usepackage{verbatim}
\usepackage{graphicx}
\usepackage{utfsym}
\usepackage{hyperref}
\usepackage{multirow}
\usepackage{longtable}
\usepackage{amsthm}

\journal{Neural Networks}

\begin{document}

\begin{frontmatter}



\title{Decomposition-based multi-scale transformer framework for time series anomaly detection} 

\author[label1]{Wenxin Zhang}
\affiliation[label1]{organization={School of Computer Science and Technology, University of Chinese Academy of Sciences},
             city={Beijing},
             postcode={100000},
             country={China}}

\author[label2]{Cuicui Luo} 

\affiliation{organization={International College, University of Chinese Academy of Sciences},
            city={Beijing},
            postcode={100000}, 
            country={China}}

\begin{abstract}
Time series anomaly detection is crucial for maintaining stable systems. Existing methods face two main challenges. First, it is difficult to directly model the dependencies of diverse and complex patterns within the sequences. Second, many methods that optimize parameters using mean squared error struggle with noise in the time series, leading to performance deterioration. To address these challenges, we propose a transformer-based framework built on decomposition (TransDe) for multivariate time series anomaly detection. The key idea is to combine the strengths of time series decomposition and transformers to effectively learn the complex patterns in normal time series data. A multi-scale patch-based transformer architecture is proposed to exploit the representative dependencies of each decomposed component of the time series. Furthermore, a contrastive learn paradigm based on patch operation is proposed, which leverages KL divergence to align the positive pairs, namely the pure representations of normal patterns between different patch-level views. A novel asynchronous loss function with a stop-gradient strategy is further introduced to enhance the performance of TransDe effectively. It can avoid time-consuming and labor-intensive computation costs in the optimization process. Extensive experiments on five public datasets are conducted and TransDe shows superiority compared with twelve baselines in terms of F1 score. Our code is available at \url{https://github.com/shaieesss/TransDe}.
\end{abstract}




\end{frontmatter}



\section{Introduction}
\label{Introduction}
With the development of the Internet of Things, an increasing number of devices and sensors have been deployed as components of real-world systems. Multivariate, massive, and complex time series data stemming from these devices brings tremendous practical significance for checking the status of equipment and maintaining the stability of systems. An elaborate time series anomaly detection algorithm can provide timely alarms for potential and disastrous malfunctions of equipment and systems, thereby sending opportune feedback to the operators and helping systems eliminate underlying hazards. Accordingly, anomaly detection for time series has momentous practical implications, which are worthy of further exploration and studies.

Some research about study univariate time series anomaly detection problem \citep{20223412588820, 20224413050293}, which is, however, unrealistic and irrational in practical life since time series are often generated from multiple devices with different functions and sensors. For example, in water distribution systems, manifold monitoring equipment reads all statuses simultaneously, such as valve status, flowing meter, and water pressure \citep{GTA}. These data with highly complex topological structure correlation provide a more comprehensive perspective of the system's state but exacerbate the difficulty of anomaly detection. Traditional algorithms fail to model representative features for time series anomaly detection because of the incapacity to deal with high-dimensional data and capture nonlinear relations and complex dependencies in sequences \citep{WOS:000883752300010}. 

Recently, deep learning methods have achieved gratifying accomplishments in time series anomaly detection problems due to their mighty power for modeling intricate representations, which mainly includes prediction-based approaches \citep{WOS:000954333200001} and reconstruction-based approaches \citep{WOS:000965416500001}. Prediction-based approaches identify anomalies by predicting future samples based on current input sequences and discriminate anomalies based on the gap between ground-truth value and the predicted observation. Reconstruction-based approaches aim to train a model to accurately reconstruct normal samples, and the samples with large reconstruction errors will be regarded as anomalies \citep{NormFAAE}. These works have shown their effectiveness and great performance on anomaly detection problems. However, research on this subject still has room for exploration. 

On the one hand, directly modeling time series can be challenging because key characteristics are often difficult to identify. This difficulty arises from the complex dependencies that exist within time series data, which typically exhibit diverse patterns such as trends, residuals, and cycles. Understanding these different patterns is essential for accurately identifying unique features, thereby improving the performance of tasks like classification, anomaly detection, and prediction. However, due to various nonlinear factors, it can be difficult for models to capture the underlying dependencies of these patterns when analyzing the original time series, complicating the modeling process. A decomposition-based approach is particularly useful in this context. Decomposition enables us to break down the time series into its main components, such as trend, seasonality, and noise. By isolating these elements, we can derive more representative latent features that facilitate clearer modeling and analysis. This strategy is especially beneficial for capturing subtle variations and fluctuations that might be lost when looking when considering the data as a whole \citep{fedformer, chen2022learning}. Furthermore, this approach simplifies the process of anomaly detection by providing the model with deeper insights for identifying anomalies based on learned characteristics.

On the other hand, most existing methods focus much sight on mean squared error(MSE) which is overly dependent on the true values in the sequence. In reality, there inevitably and ubiquitously exists noise in sequences \citep{WOS:001049240600001}, resulting in accurate prediction or reconstruction is quite difficult for models. The interference in the time series can bring large errors to MSE, which makes models fail to concentrate on analyzing normal time series. As a result, much misinformation about noise and interference is injected into the optimization process and may cause over-fitting problems. Contrastive learning has attracted much attention since its ingenious design and prominent performance, which can be an effective way of learning discriminative representations. One of the advantages of contrastive learning is to model the characteristics based on the input data, avoiding the dependency on ground truth value. Also, noise existing in different contrastive views can be effectively learned as intrinsic properties of time series instead of being regarded as deviation used for optimization.

To moderate the aforementioned challenges, we propose a novel transformer framework based on decomposition (TransDe) for multivariate time series anomaly detection. Specifically, we first decompose the original time series into trend and cyclical components to discover more representative information for contrastive learning. Then we leverage transformer architecture to learn the latent dependencies of trend and cyclical components respectively based on patch operation, generating inter-patch and intra-patch representations. Next, we take dimension expansion and fuse the learned representation of both trend and cyclical components according to patch level. Subsequently, we leverage contrastive learning to model the dependencies between inter-patch representations and intra-patch representations in normal sequences. Finally, we implement anomaly detection based on anomaly scores generated from the well-trained model.

The proposed method addresses the challenges of time series anomaly detection in several ways. First, we leverage a decomposition strategy to extract more informative components from the sequences instead of applying contrastive learning directly to the original time series. This approach allows us to capture distinct characteristics of complex time series data, reducing the difficulties associated with modeling intricate sequences and managing multiplex dependencies, ultimately enhancing the model's performance and robustness. For the second challenge, we develop an advanced contrastive representation learning paradigm for dependency, rather than relying solely on mean squared error (MSE) to assess differences from ground-truth values. By employing KL divergence as a loss computation tool, we can measure consistent information between different contrastive views while minimizing sensitivity to noise within the sequences. The asymmetric nature of KL divergence further allows for a more accurate representation of information flow and loss between distributions, helping to mitigate the effects of outliers and misinformation during the optimization process.

The main contributions of this paper are as follows:
\begin{itemize}
\item TransDe, a novel transformer framework for multivariate time series anomaly detection is proposed. It adopts a decomposition strategy to derive representative components of time series, simplifying the complexity of learning and modeling tasks for anomaly detection problems. TransDe generates independent multi-scale channels based on different patch sizes and incorporates a transformer architecture to encode dependencies from both intra-patch and inter-patch views, enabling it to capture more distinct features in normal time series.
\item A new contrastive learning approach is developed to capture distinct features in time series data. It uses KL divergence to measure consistent information between positive samples in patch-based views, reducing noise interference and managing the computational complexity typically caused by negative sample generation.
\item An asynchronous contrastive loss function with a stop-gradient strategy is proposed to prevent model collapse and strengthen the model’s learning capability.
\item Extensive comparative experiments on five public datasets and twelve baseline methods confirm the superiority of TransDe over existing methods.
\end{itemize}

The paper is structured as follows: Section \ref{Related work} reviews related works about time series anomaly detection with different algorithms and models. Section \ref{Problem statement} illustrates the definition of time series anomaly detection problems. Section \ref{Methodology} introduces the proposed model in detail. Extensive experiments and corresponding analysis are executed in Section \ref{Experiments}. Finally, Section \ref{Conclusion} concludes the whole study.

\section{Related work}
\label{Related work}
\subsection{Time series anomaly detection}
Time series anomaly detection is a prevalent research problem for its practical implications in industrial and many other fields and has been well studied. The approaches for time series anomaly detection can be divided into three categories, namely statistical methods, traditional machine learning methods, and deep learning methods \citep{20223012388180}.

Statistical methods contain auto-regressive \citep{WOS:000513451700006}, ARIMA \citep{WOS:000369269600001}, EWMA \citep{WOS:000999620800042} and so on. For example, \cite{WOS:000366879800169} develop a SARMA-based EWMA monitoring complex production system.  \cite{WOS:001146640600001} introduce the EWMA control chart to identify early neurological deterioration of ischemic stroke patients and achieve the lowest false alarm rate and the highest accuracy. Statistical methods provide an interpretable approach to data analysis and can reduce computation complexity through prior experience, but the premise for using statistical models is relatively strict and these models usually lack robustness.

Many traditional machine learning methods are adopted to detect time series anomalies, such as SVM-based \citep{WOS:000815676900032}, decision tree-based \citep{WOS:001176393100001}, clustering-based \citep{WOS:001097160000025} and so on. For example, RTtsSVM-AD \citep{RTtsSVM-AD} is introduced to detect dynamically true gait anomaly occurrence. SIM-AD \citep{SIM-AD} is a clustering-based semi-supervised approach that can operate anomaly detection tasks without human intervention. These machine learning methods have a favorable capacity towards high-dimensional time series data, but they fail to capture complex latent nonlinear dependencies among sequences.

Deep learning has been widely deployed for time series anomaly detection problems for its powerful modeling capacity towards nonlinear dependencies. Many different deep learning models are designed to tackle time series anomaly detection problems, mainly including RNN-based \citep{WOS:001098694300001}, GAN-based \citep{WOS:000965072800001}, transformer-based \citep{20231213775905}.

RNN-based methods can leverage memory units to preserve historical information. 
\cite{WOS:000964040900001} develop an RNN variational autoencoder based on inductive conformal anomaly detection to learn both spatial and temporal features of the normal dynamic behavior of the system.  \cite{WOS:000965416500001} develop an LSTM-based autoencoder to monitor indoor air quality. CSHN \citep{WOS:001096425100001} is a GRU-based cost-sensitive hybrid network to address the detection inaccuracy problem caused by the minority class of anomalies. 

GAN-based approaches train the input sequences through the dynamic adversarial process. M3GAN \citep{WOS:000991275600001} is a GAN-based approach with a masking strategy to enhance the robustness and generalization of time series anomaly detection. STAD-GAN \citep{WOS:000968706500011} improves performance on the training data containing anomalies using a self-training teacher-student framework. GRAND \citep{GRAND} integrates a variational autoencoder with a generative adversarial network to extract anomalies in the execution trace.

Transformer-based methods leverage the self-attention mechanism to encode the input time series and capture latent dependency information among the sequences. PAFormer \citep{PAFormer} is an end-to-end unsupervised parallel-attention transformer framework that is comprised of a global enhanced representation module and the local perception module. VVT \citep{VVT} leverages a transformers-based framework to effectively model the temporal dependencies and relationships among variables. STAT \citep{STAT} model semantic spatial-temporal information through spatial-temporal and temporal-spatial channels. 

Our literature review reveals that there has been significant research on time series anomaly detection, particularly using basic models, with a strong emphasis on transformer frameworks. Some studies focus on modeling the dependencies of time series directly from the original sequences. However, due to the complexity of these dependencies and the impact of nonlinear factors, it can be challenging for models to extract meaningful characteristics directly from the original data. Additionally, many models identify anomalies based on reconstruction or prediction, but their effectiveness can be hindered by large errors resulting from small anomalies. To address these challenges and capture more representative information from the sequences, we propose using a decomposition approach. This method allows us to isolate distinct components of the time series and apply a contrastive learning framework for more effective anomaly detection.

\subsubsection{Contrastive learning}
As an effective paradigm for unsupervised learning, contrastive learning aims to extract discriminative representations and common information from different views of samples, which detaches from the constraints of labels. The outset of contrastive learning is InstDic \citep{InstDic}, and later contrastive learning is extensively applied in many academic and engineering domains. Traditional contrastive learning needs to manually generate positive and negative samples, then lessen the gap between samples in the same categories and magnify the distance between samples in different categories, such as InfoNCE \citep{InfoNCE}, RINCE \citep{RINCE}, ReCo \citep{ReCo}. Unlike the traditional contrastive learning that leverages both positive and negative samples to calculate the loss function, SimSiam \citep{SimSiam} and BYOL \citep{BYOL, BYOL2} only use positive samples to compute the loss function and demonstrate its great efficiency and effectiveness.

In our study, we focus on normal time series to extract their behavioral patterns. We consider the latent representations from two different views of these sequences as positive samples. By calculating the loss based on these representations, we aim to identify anomalies that do not align with the consistent patterns observed in both views.
\section{Problem statement}
\label{Problem statement}
Consider a multivariate time series $\mathcal{X} = \{x_1, x_2, \cdots, x_T\}$, where $T$ is the length of time series. Each point $x_t \in \mathbb{R}^d$ represents the values of sensors at timestamp $t$, and $d$ is the number of sensors. We apply unsupervised learning to our proposed model with the given normal multivariate time sequence $\mathcal{X}$. Subsequently, we calculate the anomaly score $\mathcal{Y}_{test} = \{y_1, y_2, \cdots, y_{T'}\}$ for another unknown time sequence $\mathcal{X}_{test} = \{x_1, x_2, \cdots, x_{T'}\}$, where $y_t = 1$ indicates the presence of an anomaly, and $y_t = 0$ signifies normal conditions. To achieve this goal, we train a deep learning framework to understand and model the normal patterns from training samples. We then use this knowledge to detect anomalies in the test data, which includes both normal and abnormal observations.

\section{Methodology}
\label{Methodology}
In this section, we first introduce the framework of the proposed TransDe and then explain each component in detail.
\subsection{Framework}
\begin{figure*}[!htbp]
	\centering
	\includegraphics[width=1\linewidth]{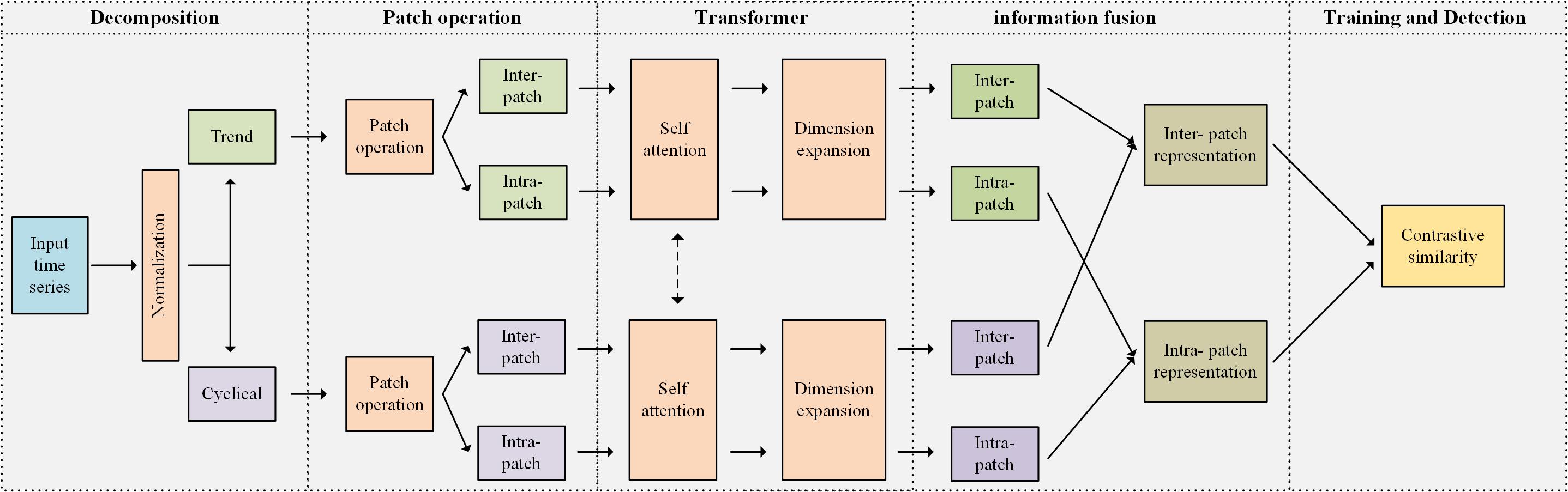}
	\caption{Overview of TransDe. The model consists of five parts, (1) sequences decomposition, (2) patch operation, (3) transformer-based representation learning, (4) dimension expansion and information fusion, and (5) training with contrastive loss and anomaly detection.}
	\label{framework}
\end{figure*}

The pipeline of TransDe is illustrated in Fig. \ref{framework}. First, the input time series are normalized and decomposed into trend and cyclical components to create more informative sequence representations. Next, each component is segmented into multiple patches of varying lengths based on different patch size scales, resulting in a multi-scale patch-based representation. A transformer with shared weights is then used to generate two contrastive views: inter-patch and intra-patch representations, which are expanded to maintain dimensional consistency. Subsequently, these multi-scale representations of these two components are concatenated according to different patch levels. Finally, a contrastive loss is obtained by calculating the similarity of two patch-level representations for time series anomaly detection.
\subsection{Decomposition}
To better capture and model the pattern information of normal sequences, we first leverage a decomposition strategy to acquire the corresponding trend components and the cyclical components of the data. There are several effective decomposition methods available for time series analysis. In this study, we have chosen to utilize the Hodrick-Prescott (HP) filter \citep{TFAD} due to its efficiency and lightweight nature. This filter allows us to effectively extract both the trend and cyclical components of the time series, facilitating a clearer understanding of the underlying patterns. The trend component reflects the underlying general characteristics of the data, helping to learn refined dependency relationships, while the cyclical component captures individual variations over time. The HP filter decomposition strategy helps reduce the complexity of modeling the original sequences and improves the ability to capture key features of different time series, which is crucial for modeling normal patterns. Additionally, this approach enhances anomaly detection, as different anomalies tend to manifest differently across the trend and cyclical components.

Specifically, denote $x_t = \tau_t + \varepsilon_t, t=(1, \cdots, T)$ as the value of time series $\mathcal{X}$ at timestamp $t$, where $\tau_t$ and $\varepsilon_t$ are respectively trend and cyclical components. The trend component $\tau_t$ can be calculated by the following optimization problem:
\begin{equation}
\label{Decomposition}
    min_\tau \quad  \sum_{t=1}^{T}(x_t - \tau_t)^2 + \alpha \sum_{t=2}^{T-1}[(\tau_{t+1} - \tau_t) - (\tau_t - \tau_{t-1})]^2
\end{equation}
where $\alpha$ is the weight parameter to measure the sensitivity of the trend to residual. 

While it is challenging to assert that the HP filter is the superior decomposition strategy, given that each method has its strengths and weaknesses, the improvements observed are significant. We will continue to build upon this work in future research.
\subsection{Patch-based attention encoder}
We develop an encoder for representation learning of trend and cyclical sequences based on patch operation, which has demonstrated strong effectiveness in representation learning \citep{DCdetector}. This patch operation can be implemented through a commonly used sliding window technique \citep{ANOMALYTRANSFORMER}.

\begin{figure}
    \centering
    \includegraphics[width=0.95\linewidth]{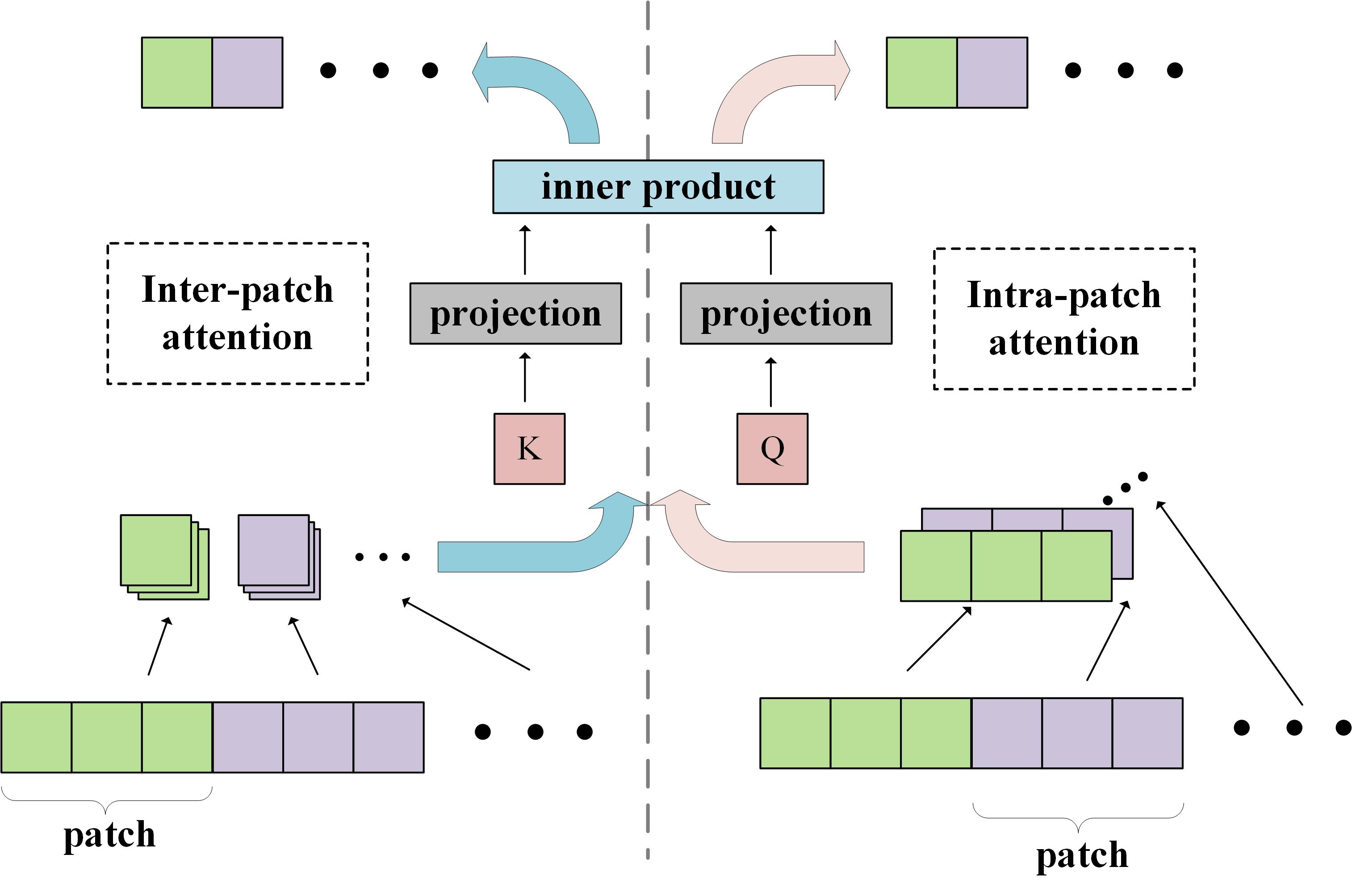}
    \caption{Attention mechanism for inter-patch and intra-patch views}
    \label{patchattention}
\end{figure}

Here’s a revised version of your text to improve accessibility for a wider audience while maintaining academic integrity:

Specifically, the input sequences $\mathcal{X} \in \mathbb{R}^{T \times d}$ are transformed into the format $\mathcal{X} \in \mathbb{R}^{P \times N \times d}$, where $P$ represents the size of each patch and $N$ denotes the number of patches. To facilitate the representation of different patch-based views,  we integrate the batch dimension with the channel dimension, resulting in the transformation $\mathcal{X}^{P \times N}$. Following this, TransDe learns attention representations for both intra-patch and inter-patch views through $L$ attention layers, as shown in Fig. \ref{patchattention}. For simplicity, the following descriptions will focus on one layer as an example.

To capture fine-grained representations, we deploy multi-scale patch operations. In practice, we set diverse patch channels with different patch sizes, which are independent of each other. The input sequences are divided into varying sizes to model dependencies at different scales. By utilizing distinct patch sizes across these independent channels, we can obtain multi-scale representations, leading to more accurate modeling outcomes.

Transformers have achieved remarkable success in time series analysis due to their exceptional ability to model global representations and capture long-term dependencies. The self-attention mechanism within transformers allows for the effective learning of spatiotemporal relationships between variables, which is crucial for modeling the normal patterns of time series and identifying anomalies. Given these advantages, we adopt transformers as the foundational architecture for our encoder to generate two contrastive views, ensuring robust anomaly detection. For intra-patch representation learning, a transformed-based framework with a multi-head attention mechanism is employed to obtain the latent dependency representations between different patches. Specifically, first, we leverage the convolutional neural network to embed patch size dimension from $N$ to $d_{model}$, obtaining the intra-patch embeddings $\mathcal{X}_{intra} \in \mathbb{R}^{P \times d_{model}}$. For each head, we initialize the transformer parameters for intra-patch representation through:
\begin{equation}
\label{intra-patch}
   Q_{intra} = W_Q \mathcal{X}_{intra}, \quad K_{intra} = W_K \mathcal{X}_{intra}
\end{equation}
where $W_Q, W_K \in \mathbb{R}^{\frac{d_{model}}{h}\times \frac{d_{model}}{h}}$ represent the query and key parameter matrices and $Q_{intra}, K_{intra} \in \mathbb{R}^{N \times \frac{d_{model}}{h}}$ are the corresponding embeddings. Next, the normalized attention dependency representation can be derived:
\begin{equation}
\label{intra-patch softmax}
   Z_{intra} = \sigma(\frac{Q_{intra} \cdot K_{intra}'}{\sqrt{\frac{d_{model}}{h}}})
\end{equation}
Last, we concatenate the obtained representation $\Omega_{intra} \in \mathbb{R}^{P \times P}$ of different head:
\begin{equation}
\label{intra-patch cat}
   \Omega_{intra} = cat(Z_{intra}^1, \cdots, Z_{intra}^h)
\end{equation}
where $cat(\cdot)$ is concatenation function and $h$ is the number of heads.

Similarly, for inter-patch representation learning, we derive the inter-patch embeddings $\mathcal{X}_{inter} \in \mathbb{R}^{N \times d_{model}}$ by embedding patch number dimension from $P$ to $d_{model}$, then we initialize the transformer parameters for inter-patch representation through:
\begin{equation}
\label{inter-patch}
   Q_{inter} = W_Q \mathcal{X}_{inter}, \quad K_{inter} = W_K \mathcal{X}_{inter}
\end{equation}
where $W_Q, W_K \in \mathbb{R}^{\frac{d_{model}}{h}\times \frac{d_{model}}{h}}$ represent the query and key parameter matrices and $Q_{inter}, K_{inter} \in \mathbb{R}^{N \times \frac{d_{model}}{h}}$ are the corresponding embeddings. Next, the normalized attention dependency representation can be derived:
\begin{equation}
\label{inter-patch softmax}
   Z_{inter} = \sigma(\frac{Q_{inter} \cdot K_{inter}'}{\sqrt{\frac{d_{model}}{h}}})
\end{equation}
where $\sigma( \cdot)$ is softmax normalization function. Last, we concatenate the obtained representation $\Omega_{inter} \in \mathbb{R}^{N \times N}$ of different head:
\begin{equation}
\label{inter-patch cat}
   \Omega_{inter} = cat(Z_{inter}^1, \cdots, Z_{inter}^h)
\end{equation}

We leverage shared matrices $W_Q, W_K$ to learn the patch-based attention for two components to keep the consistency of latent vector space.

\subsection{Dimension expansion and information fusion}
\begin{figure}
    \centering
    \includegraphics[width=0.5\linewidth]{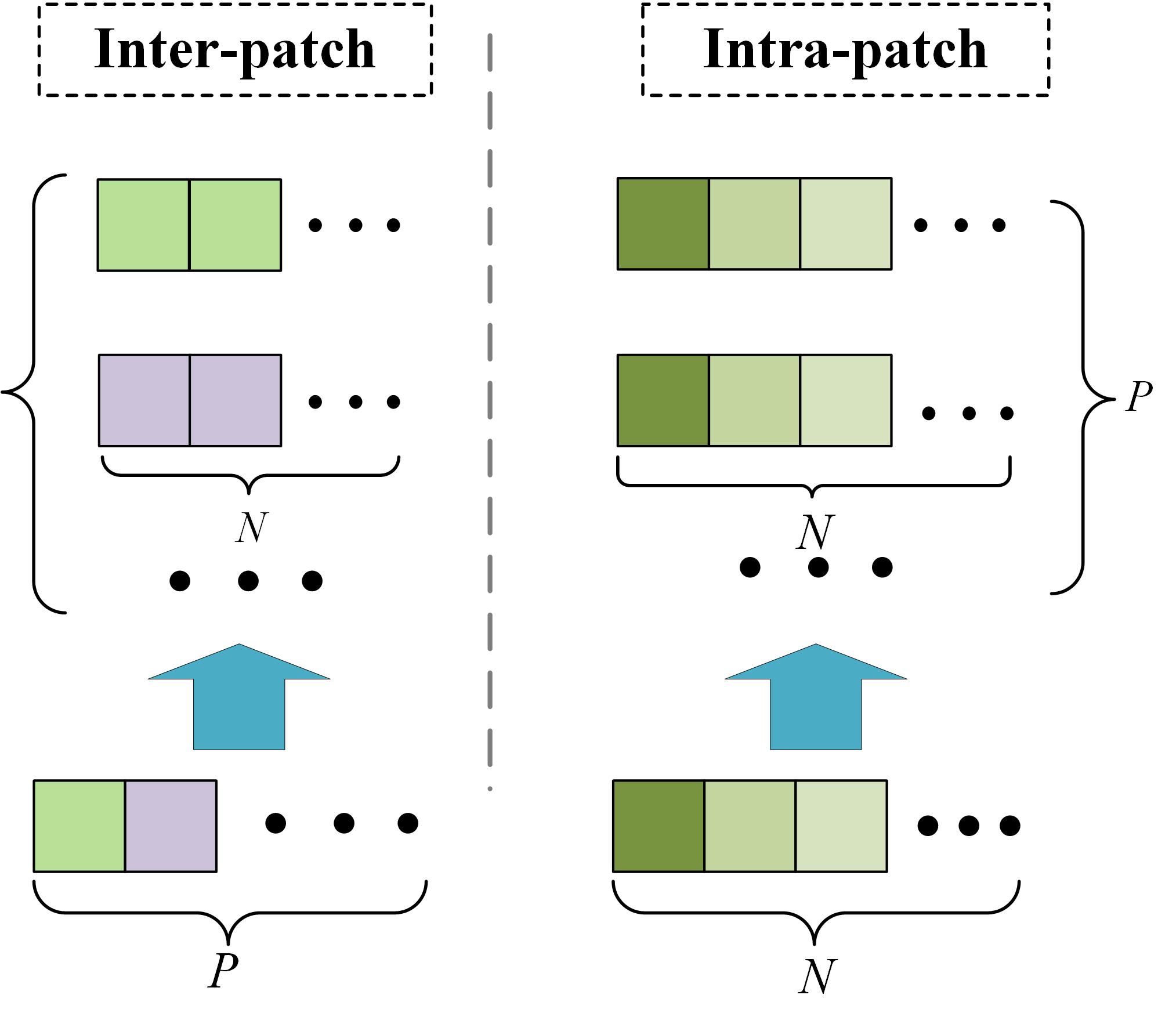}
    \caption{A simple example of dimension expansion for different patch views}
    \label{Dimensionexpansion}
\end{figure}

After obtaining patch-based dependency information for trend and cyclical sequences, we need to integrate these representations for more effective contrastive learning. However, the encoded intra-patch dependency representation $\Omega_{intra}$ and the inter-patch dependency representation $\Omega_{inter}$ have different matrix dimensions, making direct calculations challenging. To overcome this issue, we adopt dimension expansion to achieve consistency in matrix dimensions, as shown in Fig. \ref{Dimensionexpansion}. 

Specifically, for a given time series, the corresponding intra-patch and inter-patch dependency representations are denoted as $\Omega_{intra} \in \mathbb{R}^{1 \times P\times P}$ and $\Omega_{inter} \in \mathbb{R}^{1 \times N \times N}$, respectively. Given that $T = P \times N$ based on the patch operation, we can modify the intra-patch representation by repeating its elements $N$ times, then the intra-patch dependency representation can be transformed to $\Omega_{intra} \in \mathbb{R}^{1 \times T \times T}$.  Similarly, we can transform the inter-patch representation by repeating its elements $P$ times, yielding $\Omega_{inter} \in \mathbb{R}^{1 \times T\times T}$. 

After dimension expansion, we achieve the consistency between inter-patch dependency representation and inter-patch dependency representation. Denote the intra-patch representation and inter-patch representation respectively as $\Omega_{intra}^{tr}$ and $\Omega_{inter}^{tr}$ for trend component, and $\Omega_{intra}^{cy}$ and $\Omega_{inter}^{cy}$ for cyclical component. Then we implement information fusion to generate composite embeddings:
\begin{equation}
    \hat{\Omega}_{intra} = cat(\Omega_{intra}^{tr}, \Omega_{intra}^{cy})
\end{equation}
\begin{equation}
    \hat{\Omega}_{inter} = cat(\Omega_{inter}^{tr}, \Omega_{inter}^{cy})
\end{equation}
where $\hat{\Omega}_{intra}$ and $\hat{\Omega}_{inter}$ are respectively fused intra-patch dependency embeddings and inter-patch dependency embeddings.

\subsection{Discrepancy contrastive learning}
Patch-based attention encoders have a great capacity for modeling local information of sequences. However, different attention strategies focus on dependencies at varying scales. Specifically, intra-patch representation learning captures the relationships among elements within a single patch but overlooks the relationships between different patches. On the other hand, inter-patch representation learning focuses on relationships between patches while neglecting the internal dependencies within each patch. To account for both scales, we treat $\hat{\Omega}_{intra}$ and $\hat{\Omega}_{inter}$ as two distinct views for contrastive learning. This approach helps us capture representative knowledge of normal patterns in time series that are absent in abnormal patterns.

To implement this, we define a Kullback-Leibler (KL) divergence loss function to calculate the similarity between $\hat{\Omega}_{intra}$ and $\hat{\Omega}_{inter}$. In addition, we use a stop-gradient operation for asynchronous training, which has proven effective in preventing model collapse \citep{SimSiam}. The loss function for $\hat{\Omega}_{intra}$ and $\hat{\Omega}_{inter}$ can be respectively formulated as follows:
\begin{equation}
\label{Lintra}
    \mathcal{L}_{intra} = \sum KL(\hat{\Omega}_{intra}, Stop(\hat{\Omega}_{inter})) + KL(Stop(\hat{\Omega}_{inter}), \hat{\Omega}_{intra})
\end{equation}
\begin{equation}
\label{Linter}
    \mathcal{L}_{inter} = \sum KL(\hat{\Omega}_{inter}, Stop(\hat{\Omega}_{intra})) + KL(Stop(\hat{\Omega}_{intra}), \hat{\Omega}_{inter})
\end{equation}
where $KL(\cdot || \cdot)$ is KL divergence and $Stop(\cdot)$ denotes stop gradient operation. We define different patch size as different channels, then the overall loss function can be formulated as:
\begin{equation}
    \mathcal{L} = \frac{\mathcal{L}_{intra} - \mathcal{L}_{inter}}{C}
\end{equation}
where $C$ is the number of channels.

Unlike traditional contrastive learning methods that use both positive and negative samples to define category characteristics, our approach focuses solely on intra-patch and inter-patch views as positive pairs for loss calculation. The typical use of negative samples, taken from other examples in the same batch, presents significant challenges. This method is time-consuming because it requires extensive calculations of similarity to select high-quality negative samples, which increases the complexity of the process. Additionally, the imbalance in samples often found in anomaly detection problems worsens this issue; the quality of samples from minority classes often differs greatly from that of majority class samples, leading to uneven learning efficiencies for different labels. Therefore, by strategically ignoring negative samples, especially those that are far away or less informative, we improve the efficiency of the optimization algorithm and enhance the model’s learning ability. This strategy helps remove redundant and potentially confusing information that could hinder the training process.

Empirical evidence in Section 5.6 demonstrates that our approach of ignoring negative samples can achieve performance comparable to, or even better than, established methods like SimCLR and MoCo, which utilize techniques to select a smaller set of negative samples.

Thus, the intentional exclusion of negative samples in contrastive learning is a thoughtful approach that balances computational efficiency with learning effectiveness, leading to significant improvements in both training speed and model accuracy. This method aligns with the foundational principles of contrastive learning and is supported by recent research findings.

\subsection{Anomaly detection}
During the testing phase, test sequences are directly input into TransDe. Since TransDe is trained solely on normal data, any anomalies with rare patterns will result in inconsistent representations, leading to a high anomaly score. We utilize the similarity of the encoded representations as our anomaly score, which can be expressed as follows:
\begin{equation}
    s(\mathcal{X}) = \sum [KL(\hat{\Omega}_{intra}, Stop(\hat{\Omega}_{inter})) + KL(\hat{\Omega}_{inter}, Stop(\hat{\Omega}_{intra}))]
\end{equation}

To facilitate anomaly detection, we set a hyperparameter threshold $\rho$. If the calculated score exceeds this threshold, the output is classified as an anomaly. This can be formulated as follows:
\begin{equation}
    \mathcal{Y} = 
    \begin{cases}
        1, \quad s(\mathcal{X}) \geq \rho\\
        0, \quad s(\mathcal{X}) < \rho
    \end{cases}
\end{equation}
\section{Experiments}
\label{Experiments}
In this section, we carry out extensive experiments using five public datasets to verify the effectiveness of TransDe and its superiority over baselines.
\subsection{Datasets}
The experiments are conducted using the following real-world datasets. A detailed statistical description is provided in Table \ref{DATASETS}. In this table, dimension represents the number of variables in the original datasets, while continuous variables and discrete variables denote the number of variables with different data types. Training and testing specify the number of samples used for the respective training and testing processes, and anomaly rate denotes the proportion of anomalies relative to the total number of samples.
\textbf{MSL} \citep{MSLandSMAP}: The Mars Science Laboratory (MSL) dataset, collected by NASA, contains operational data from the sensors and actuators of the Mars rover.

\textbf{SMAP} \citep{MSLandSMAP}: The Soil Moisture Active Passive (SMAP) dataset monitors soil moisture on Earth using satellite observations. It records data every two to three days, enabling the consideration of both climate changes and seasonal variations.

\textbf{PSM} \citep{PSM}: The Pooled Server Metric (PSM) dataset is collected from server machines at eBay.

\textbf{SMD}  \citep{SMD}: The Server Machine Dataset (SMD) consists of data from 28 different machines, divided into 28 subsets for individual training and testing.

\textbf{SWaT} \citep{SWaT}: The Secure Water Treatment (SWaT) dataset is a publicly available time series benchmark derived from infrastructure systems. It involves collecting raw water, processing it with essential chemical agents, and purifying it through ultrafiltration and ultraviolet radiation.
\begin{table*}[!htbp]
	\centering
	\caption{Statistics of datasets}
	\label{DATASETS}
 \resizebox{\textwidth}{!}{
	\begin{tabular}{c c c c c c c c} 
		\hline
		Dataset & Application & Dimension & Continuous variables & Discrete variables & Training & Testing & Anomaly rate(\%)\\\hline
		MSL & Space & 55 & 1 & 54 & 58317 & 73729 & 5.54\\
		SMAP & Space & 25 & 1 & 24 & 135183 & 427617 & 13.13\\
		PSM & Server & 26 & 25 & 0 & 132481 & 87841 & 27.76\\
		SMD & Server & 38 & 37 & 1 & 708405 & 708420 & 4.16\\
		SWaT & Water & 51 & 25 & 26 & 495000 & 449919 & 11.98\\\hline
	\end{tabular}
 }
\end{table*}
\subsection{Baselines}
We compare TransDe with twelve baselines to demonstrate the effectiveness of TransDe. Wherein, LOF, IForest, and DAGMM are classic anomaly detection approaches, the rest can be categorized as deep learning methods. The details of baselines are shown in Table \ref{baselines}.

\textbf{LOF} \citep{LOF}. It is an anomaly detection approach used for outlier detection based on local density deviation.

\textbf{IForest} \citep{IForest}. It is an assembled framework that identifies anomalies through a stochastic selection of features.

\textbf{DAGMM} \citep{DAGMM}. It is a density-based anomaly detection method that leverages the Gaussian mixture model to reconstruct normal sequences.

\textbf{OmniAnomaly} \citep{OmniAnomaly}. It is a reconstruction-based stochastic recurrent neural network(RNN) framework for multivariate time series anomaly detection.

\textbf{InterFusion} \citep{InterFusion}. It is an unsupervised learning approach that leverages a hierarchical variational auto-encoder to fit the normal patterns of input sequences. 

\textbf{TS-CP2} \citep{TS-CP2}. It is a contrastive learning-based anomaly detection method integrated with change point detection.

\textbf{TranAD} \citep{TranAD}. It is a deep transformer diagnosis model that utilizes attention-based encoders to rapidly learn knowledge of temporal dependencies in sequences and carry out inference.

\textbf{AnomalyTrans} \citep{AnomalyTrans}. It is a transformer-based anomaly detection method that amplifies the normal-abnormal distinguishability of the association discrepancy through a minimax strategy.

\textbf{DCFF-MTAD} \citep{DCFF-MTAD}. It is a dual-channel multivariate time-series anomaly detection approach that employs spatial short-time Fourier transform and a graph attention network.

\textbf{MAUT} \citep{MAUT}. It is a U-shaped transformer framework that leverages memory units to learn behavior patterns of normal sequences.

\textbf{ATF-UAD} \citep{ATF-UAD}. It is a reconstruction-based anomaly detection approach that decomposes time series into time and frequency domains.

\textbf{NormFAAE} \citep{NormFAAE} It is a backbone end-to-end learning framework for anomaly detection problems that integrates two learnable normalization sub-modules with a dual-phase filter-augmented auto-encoder to reconstruct the normal data.

\begin{table}[!htbp]
	\centering
	\caption{Details of baselines}
	\label{baselines}
	\begin{tabular}{c | c c c c} 
	\hline
	Model & Publication & year & Authors \\\hline
	LOF & ACM SIGMOD & 2000 & \citep{LOF} \\
	IForest & ACM TKDD & 2012 & \citep{IForest} \\
	DAGMM & ICLR & 2018 & \citep{DAGMM} \\
        OmniAnomaly & ACM SIGKDD & 2019 & \citep{OmniAnomaly} \\
        InterFusion & ACM SIGKDD & 2021 & \citep{InterFusion} \\
        TS-CP2 & ACM WWW & 2021 & \citep{TS-CP2} \\
        TranAD & VLDB & 2022 & \citep{TranAD} \\
        AnomalyTrans & ICLR & 2022 & \citep{AnomalyTrans} \\
	DCFF-MTAD & Sensors & 2023 & \citep{DCFF-MTAD} \\
        MAUT & IEEE ICASSP & 2023 & \citep{MAUT}\\
        ATF-UAD & Neural Networks & 2023 & \citep{ATF-UAD} \\
	NormFAAE & Neural Networks & 2024 & \citep{NormFAAE} \\\hline
	\end{tabular}
\end{table}

\subsection{Evaluation metrics}
We leverage precision(P), recall(R), and F1-score(F1) to measure the performance of extensive experiments. Precision is the ratio of true positive anomalies to all predicted anomalies. Recall is the ratio of true positive anomalies to all real anomalies. F1-score is an indicator that comprehensively considers the performance of both precision and recall. The formulation of P, R, and F1 are as follows:
\begin{equation}
\label{P}
    P = \frac{TP}{TP + FP}
\end{equation}
\begin{equation}
\label{R}
    R = \frac{TP}{TP + FN}
\end{equation}
\begin{equation}
\label{F1}
    F1 = \frac{2 \times P \times R}{P + R}
\end{equation}
where $TP$, $FP$, and $FN$ respectively indicate true positives, false positives, and false negatives.
\subsection{Implementation details}
Our experiments are implemented with PyTorch and all methods are executed in a Python 3.9.12 environment, utilizing a single NVIDIA A40 GPU, 40GB of RAM, and a 2.60GHz Xeon(R) Gold 6240 CPU.
For the baseline methods, we implement them using the source code provided by the authors.

Our TransDe includes two encoder layers. The hidden size is set to 128, and the number of attention heads is 1. We employ Adam optimizer and initialize the learning rate as $10^{-4}$. The batch size is set to 128 for all experiments. The more detailed experimental setup is shown in Table \ref{setup}. 
\begin{table}[!htbp]
	\centering
	\caption{Details of experiments on different benchmark datasets}
	\label{setup}
	\begin{tabular}{c | c c} 
		\hline
		Dataset & Window & Patch size \\\hline
		MSL & 90 & [3, 5] \\
		SMAP & 105 & [3, 5, 7] \\
		PSM & 60 & [1, 3, 5] \\
		SMD & 105 & [5, 7] \\
		SWaT & 105 & [3, 5, 7] \\\hline
	\end{tabular}
\end{table}

\subsection{Performance}
Table \ref{performance} shows the evaluation results of our model TransDe compared with twelve baselines. The best results are highlighted in Bold, and the second are underlined.

\begin{table*}[!htb]
	\centering
	\caption{Overall performances on public datasets. The P, R, and F1 respectively represent precision, recall, and F1-score. The best results are highlighted in Bold, and the second are underlined. All results are in \%.}
	\label{performance}
 \resizebox{\textwidth}{!}{
	\begin{tabular}{c | c c c | c c c | c c c | c c c | c c c }
		\hline
		Dataset & \multicolumn{3}{c|}{SMD} & \multicolumn{3}{c|}{MSL} & \multicolumn{3}{c|}{SMAP} & \multicolumn{3}{c|}{SWaT} & \multicolumn{3}{c}{PSM}\\\hline
            Metrix & P & R & F1 & P & R & F1 & P & R & F1 & P & R & F1 & P & R & F1\\\hline
            LOF & 56.34 & 39.86 & 46.68 & 47.72 & 85.25 & 61.18 & 58.93 & 56.33 & 57.60 & 72.15 & 65.43 & 68.62 & 57.89 & 90.49 & 70.61\\
            IForest & 42.31 & 73.29 & 53.64 & 53.94 & 86.54 & 66.45 & 52.39 & 59.07 & 55.53 & 49.29 & 44.95 & 47.02 & 76.09 & 0.92.45 & 83.48\\
            DAGMM & 67.30 & 49.89 & 57.30 & 89.60 & 63.93 & 74.62 & 86.45 & 56.73 & 68.51 & 89.92 & 57.84 & 70.40 & 93.49 & 70.03 & 80.08\\
            OmniAnomaly & 83.68 & 86.82 & 85.22 & 89.02 & 0.86.37 & 87.67 & 92.49 & 81.99 & 86.92 & 81.42 & 84.30 & 82.83 & 88.39 & 74.46 & 80.83\\
            InterFusion & 87.02 & 85.43 & 86.22 & 81.28 & 92.70 & 86.62 & 89.77 & 88.52 & 89.14 & 80.59 & 85.58 & 83.10 & 83.61 & 83.45 & 83.52\\
            TS-CP2 & 87.42 & 66.25 & 75.38 & 86.45 & 68.48 & 76.42 & 87.65 & 83.18 & 85.36 & 81.23 & 74.10 & 77.50 & 82.67 & 78.16 & 80.35\\
            TranAD & 95.88 & 91.72 & $\underline{91.57}$ & 90.38 & 95.78 & 93.04 & 80.43 & 99.99 & 89.15 & 97.60 & 69.97 & 81.51 & 89.51 & 89.07 & 89.29\\
            AnomalyTrans & 88.47 & 92.28 & 90.33 & 91.92 & 96.03 & 93.93 & 93.59 & 99.41 & $\underline{96.41}$ & 89.10 & 99.28 & $\underline{94.22}$ & 96.14 & 95.31 & $\underline{95.72}$\\
            DCFF-MTAD & 84.16 & 99.99 & 91.40 & 92.57 & 94.78 & 93.66 & 97.67 & 82.68 & 89.55 & 89.56 & 91.55 & 90.56 & 93.52 & 90.17 & 91.81\\
            MAUT & 86.45 & 82.54 & 84.45 & 93.99 & 94.52 & $\underline{94.25}$ & 96.12 & 95.36 & 95.74 & 96.13 & 81.43 & 88.17 & 95.49 & 88.58 & 91.91\\
            ATF-UAD & 83.12 & 81.05 & 82.07 & 91.32 & 92.56 & 91.94 & 87.50 & 41.18 & 55.99 & 99.99 & 68.79 & 81.51 & 76.74 & 93.65 & 84.36\\
            NormFAAE & 95.90 & 95.85 & $\mathbf{93.82}$ & 80.22 & 89.44 & 84.58 & 76.05 & 87.21 & 81.25 & 91.25 & 86.52 & 88.82 & 92.56 & 91.74 & 92.15\\\hline
            TransDe(Ours) & 86.64 & 88.03 & 87.33 & 92.89 & 95.87 & $\mathbf{94.36}$ & 94.35 & 99.10 & $\mathbf{96.67}$ & 97.53 & 98.46 & $\mathbf{98.04}$ & 93.13 & 98.46 & $\mathbf{96.43}$\\\hline
	\end{tabular}
 }
\end{table*}

Among the three traditional methods, LOF performs worst on the SMD, MSL, and PSM datasets, as measured by the F1 metric. In contrast, IForest underperforms on the SMAP and SWaT datasets. The poorer performance of LOF may be attributed to its focus on local neighboring information, which fails to capture broader, global patterns. Similarly, IForest’s random selection strategy can mask and overlook key characteristics of the data. In contrast, DAGMM outperforms other traditional methods on most datasets, primarily due to its strong capability to model the distribution of input sequences through a stacked linear composition. Overall, traditional methods show significant room for improvement, largely due to their limited ability to learn nonlinear relationships and model complex dependencies.

Deep learning-based baselines generally outperform traditional approaches. Specifically, TS-CP2 identifies time-dependent changes in trends using change point detection methods, leading to better performance than the three traditional methods. OmniAnomaly leverages RNN to reconstruct time sequences, while ATF-UAD rebuilds sequences by analyzing both frequency and time domains. Both methods effectively capture the trends in time series, enhancing their performance. InterFusion employs a variational auto-encoder to model the pattern of normal time series. Its robust capacity to transform and capture nonlinear representations allows it to maintain consistently high performance, achieving F1 scores above 0.8 across various datasets. Additionally, DCFF-MTAD leverages graph neural networks and Fourier transform to model the spatiotemporal information of sequences, yielding competitive results compared to some transformer-based methods.

NormFAAE, MAUT, TranAD, and AnomalyTrans are all transformer-based methods. Transformer can capture deep latent representations of time series and dependencies within time series through the self-attention mechanism. With the help of transformer architecture, these models capture deep dependencies of normal patterns among the time series, which can not be modeled in anomalies. Consequently, they achieve notable improvements in experimental performance. The average F1 scores for NormFAAE, MAUT, TranAD, and AnomalyTrans across five datasets are 0.8812, 0.9090, 0.8891, and 0.9412, respectively, indicating satisfactory results.

TransDe outperforms most models across the datasets, highlighting its effectiveness and superiority over the baseline methods. Specifically, it achieves increases in the F1 score of 0.11\%, 0.26\%, 3.82\%, and 0.71\% on the MSL, SMAP, SWaT, and PSM datasets, respectively. These improvements can be attributed to TransDe's ability to extract representative information from sequences through HP filter decomposition, which aids in identifying key factors in time series data. Additionally, the patch-level transformer encoder facilitates the learning of the various decomposed components of sequences, allowing it to capture latent dependencies within the input data. As a result, TransDe effectively learns and models the behavioral patterns of normal time series. However, it is worth noting that TransDe performs relatively poorly on the SMD dataset. This underperformance may be due to the model mistakenly classifying some normal signals as anomalies. Furthermore, the presence of numerous single-point anomalies in the SMD dataset can complicate the model's ability to identify subtle anomalies that lack distinctive features.

\subsection{Ablation study}
In this section, we conduct ablation experiments to verify the effectiveness of each component in the model. 

First, we validate the effectiveness of the stop-gradient strategy. Using Equations \ref{Linter} and \ref{Lintra}, we compute KL divergence loss with the stop-gradient applied in both the inter-patch and intra-patch branches. The results are shown in Table \ref{stopablation}. Our findings indicate that TransDe achieves optimal performance when employing both stop-gradient strategies, whereas its performance is significantly reduced in the absence of these strategies. These results highlight that stop-gradient strategies facilitate the learning of more representative embeddings and enhance the model's ability to capture latent dependencies within sequences. Furthermore, it is noteworthy that even without the stop-gradient strategies, TransDe still outperforms most baseline models, underscoring its overall effectiveness.

Second, we conduct ablation experiments on the patch-level operation. In TransDe, we process each decomposed component of original sequences at respectively inter-patch and intra-patch levels to capture local and global dependencies of time series. We compare the effectiveness of TransDe, which employs a combination of multi-scale patch levels, to a model that uses only pure patch levels. The results are presented in Table  \ref{patchablation}. Our findings reveal that a model with two inter-patch channels struggles to capture local dependencies, while a model with two intra-patch channels fails to differentiate effectively between different patches. This limitation results in relatively poor performance compared to TransDe. In contrast, the integration of both inter-patch and intra-patch channels enhances the model's ability to learn deep, distinguishing representations that incorporate both local and global information. Therefore, we conclude that both inter-patch and intra-patch branches significantly contribute to the model's robust representation learning capacity.

Third, we conduct ablation experiments on the loss function. In TransDe, we leverage symmetrical KL divergence as the loss function to measure the distance between representations at different patch levels. For comparison, we also employ simple asymmetrical KL divergence and Jensen-Shannon (JS) divergence. The results are presented in Table \ref{lossablation}. Our observations indicate that TransDe with asymmetrical KL divergence demonstrates relatively poor performance, while the use of JS divergence leads to a significant decline in performance. These findings suggest that symmetrical KL divergence is more effective in capturing discriminative information regarding latent dependencies within the sequences. Therefore, we conclude that TransDe with symmetrical KL divergence achieves the best performance compared to the other loss function variants.

\begin{table*}[!htbp]
	\centering
	\caption{The results of ablation experiments on stop gradient. The best results are in bold. All results are in \%.}
	\label{stopablation}
\resizebox{\textwidth}{!}{
	\begin{tabular}{c c | c c c | c c c | c c c | c c c | c c c }
		\hline
		\multicolumn{2}{c|}{Stop gradient} & \multicolumn{3}{c|}{SMD} & \multicolumn{3}{c|}{MSL} & \multicolumn{3}{c|}{SMAP} & \multicolumn{3}{c|}{SWaT} & \multicolumn{3}{c}{PSM}\\\hline
            Intra & Inter & P & R & F1 & P & R & F1 & P & R & F1 & P & R & F1 & P & R & F1\\\hline
            \usym{1F5F4} & \usym{1F5F4} & 80.15 & 80.06 & 80.10 & 89.26 & 90.52 & 89.88 & 93.11 & 95.69 & 94.38 & 88.54 & 90.96 & 89.73 & 91.56 & 90.14 & 90.84\\
            \usym{1F5F8} & \usym{1F5F4} & 85.01 & 85.34 & 85.17 & 91.14 & 92.45 & 91.79 & 96.83 & 93.85 & 95.31 & 96.87 & 96.98 & 96.92 & 92.24 & 95.42 & 93.80\\
            \usym{1F5F4} & \usym{1F5F8} & 86.11 & 85.36 & 85.73 & 91.43 & 95.14 & 93.25 & 93.17 & 94.20 & 93.68 & 96.05 & 92.51 & 94.25 & 91.98 & 97.52 & 94.67\\
            \usym{1F5F8} & \usym{1F5F8} & 86.64 & 88.03 & $\mathbf{87.33}$ & 92.89 & 95.87 & $\mathbf{94.36}$ & 94.35 & 99.10 & $\mathbf{96.67}$ & 97.53 & 98.46 & $\mathbf{98.04}$ & 93.13 & 98.46 & $\mathbf{96.43}$\\\hline
	\end{tabular}
}
\end{table*}
\begin{table*}[!htbp]
	\centering
	\caption{The results of ablation experiments on patch level. The best results are in bold. All results are in \%.}
	\label{patchablation}
\resizebox{\textwidth}{!}{
	\begin{tabular}{c c | c c c | c c c | c c c | c c c | c c c }
		\hline
		\multicolumn{2}{c|}{Patch level} & \multicolumn{3}{c|}{SMD} & \multicolumn{3}{c|}{MSL} & \multicolumn{3}{c|}{SMAP} & \multicolumn{3}{c|}{SWaT} & \multicolumn{3}{c}{PSM}\\\hline
            Intra & Inter & P & R & F1 & P & R & F1 & P & R & F1 & P & R & F1 & P & R & F1\\\hline
            \usym{1F5F8} & \usym{1F5F4} & 80.12 & 82.78 & 81.42 & 89.63 & 92.11 & 90.85 & 93.85 & 91.10 & 92.45 & 92.56 & 91.86 & 92.20 & 98.96 & 90.14 & 94.34\\
            \usym{1F5F4} & \usym{1F5F8} & 82.14 & 81.72 & 81.92 & 88.52 & 84.32 & 86.36 & 90.11 & 90.52 & 90.31 & 91.04 & 89.64 & 90.33 & 86.35 & 92.41 & 89.27\\
            \usym{1F5F8} & \usym{1F5F8} & 86.64 & 88.03 & $\mathbf{87.33}$ & 92.89 & 95.87 & $\mathbf{94.36}$ & 94.35 & 99.10 & $\mathbf{96.67}$ & 97.53 & 98.46 & $\mathbf{98.04}$ & 93.13 & 98.46 & $\mathbf{96.43}$\\\hline
	\end{tabular}
}
\end{table*}
\begin{table*}[!htbp]
	\centering
	\caption{The results of ablation experiments on loss function. The best results are in bold. All results are in \%.}
	\label{lossablation}
\resizebox{\textwidth}{!}{
	\begin{tabular}{c | c c c | c c c | c c c | c c c | c c c }
		\hline
		Datasets & \multicolumn{3}{c|}{SMD} & \multicolumn{3}{c|}{MSL} & \multicolumn{3}{c|}{SMAP} & \multicolumn{3}{c|}{SWaT} & \multicolumn{3}{c}{PSM}\\\hline
            Loss function & P & R & F1 & P & R & F1 & P & R & F1 & P & R & F1 & P & R & F1\\\hline
            JS & 75.21 & 78.63 & 76.88 & 85.63 & 88.54 & 87.06 & 89.31 & 85.72 & 87.48 & 86.97 & 88.29 & 87.63 & 90.52 & 84.57 & 87.44\\
            Simple KL & 85.42 & 85.98 & 85.70 & 90.56 & 94.89 & 92.67 & 93.51 & 97.99 & 95.70 & 96.42 & 97.81 & 97.11 & 91.73 & 96.98 & 94.28\\
            TransDe & 86.64 & 88.03 & $\mathbf{87.33}$ & 92.89 & 95.87 & $\mathbf{94.36}$ & 94.35 & 99.10 & $\mathbf{96.67}$ & 97.53 & 98.46 & $\mathbf{98.04}$ & 93.13 & 98.46 & $\mathbf{96.43}$\\\hline
	\end{tabular}
}
\end{table*}

\begin{figure*}[!htb]
\centering
\subfloat[The F1 metric]{
		\includegraphics[scale=0.3]{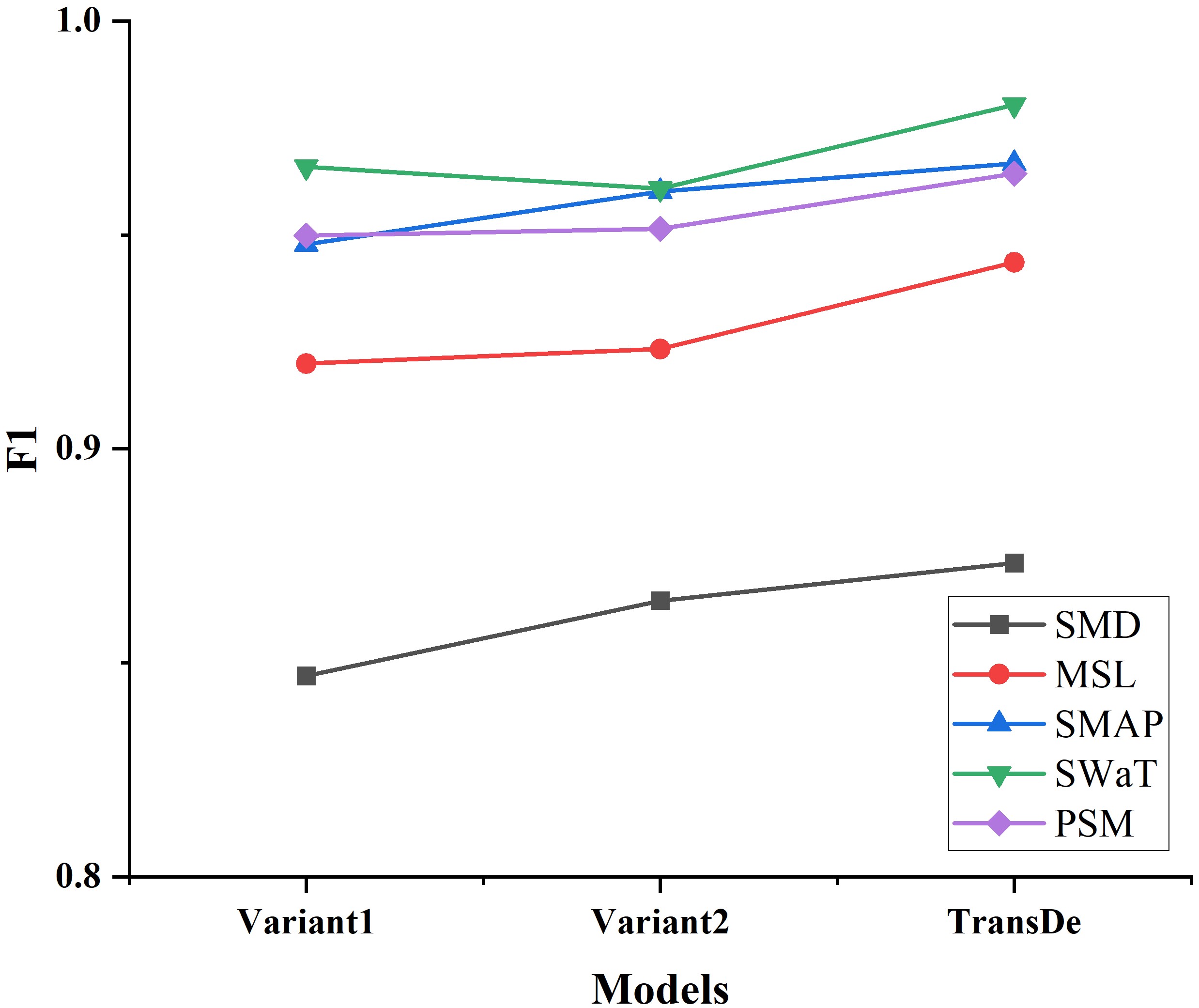} \label{f1norm}}
\subfloat[The ACC metric]{
		\includegraphics[scale=0.3]{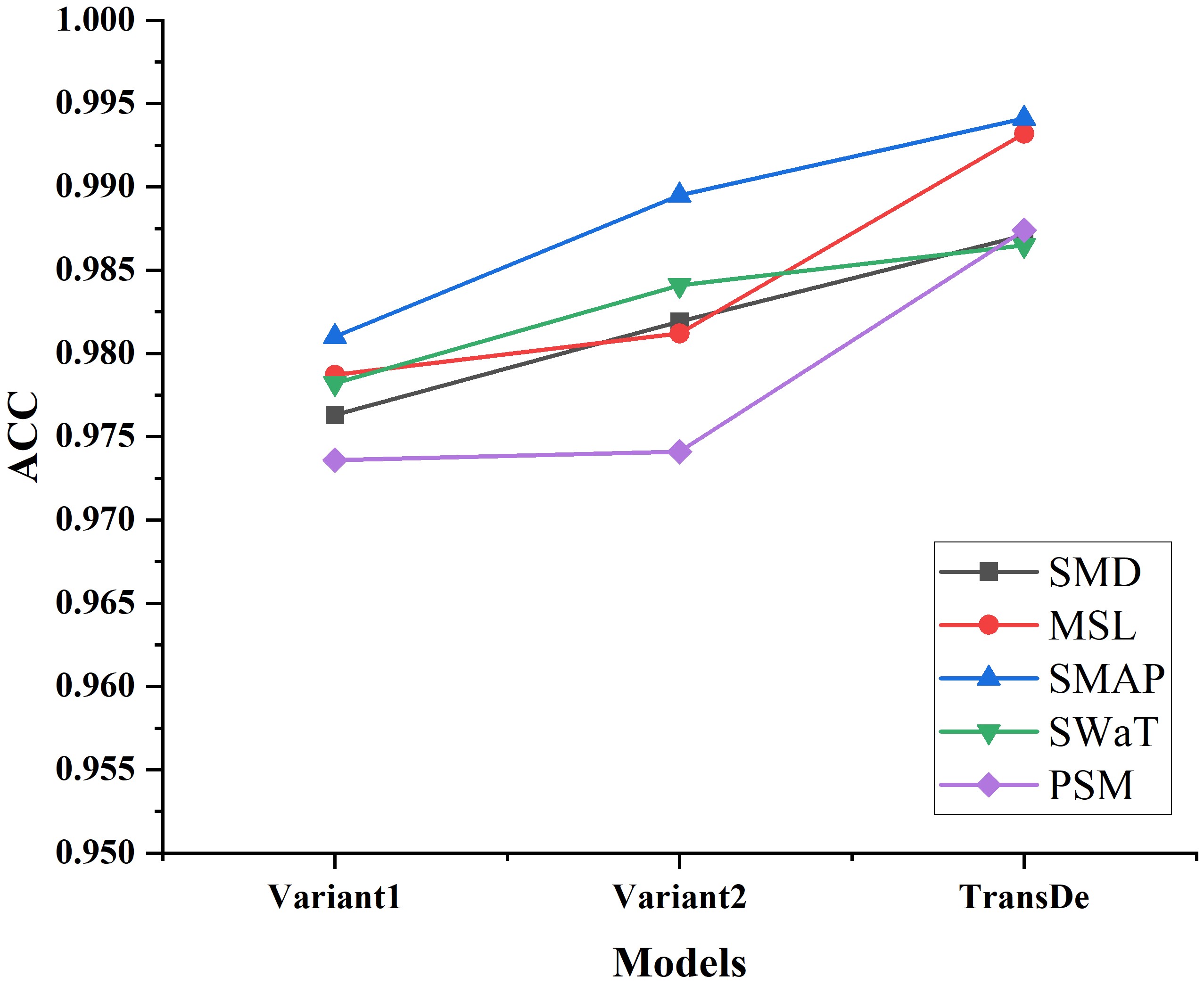} \label{accnorm}}
\caption{The ablation experiments of normalization operation}
\label{fig:enter-label}
\end{figure*}

Fourth, we examine the impact of normalization operations on the model's performance. In TransDe, we apply normalization to the original input sequences, and we explore different placements for this normalization step. The results are shown in Fig. \ref{fig:enter-label}. In this figure, Variant 1 represents TransDe without any normalization, while Variant 2 indicates TransDe with normalization applied to the two distinct components: the trend and the cyclical components.

Our observations indicate that omitting the normalization operation leads to a slight decrease in performance. This is because normalization helps limit the range of sequences, making it easier to capture standardized features. However, normalizing the original input yields better performance than normalizing the decomposed components. When different decomposed components are normalized, it may disrupt their intrinsic representative features. In summary, TransDe demonstrates the best performance compared to the two variants, highlighting the importance of normalization on the original input sequences.

\begin{figure*}[!htb]
\centering
\subfloat[The F1 metric]{
		\includegraphics[scale=0.26]{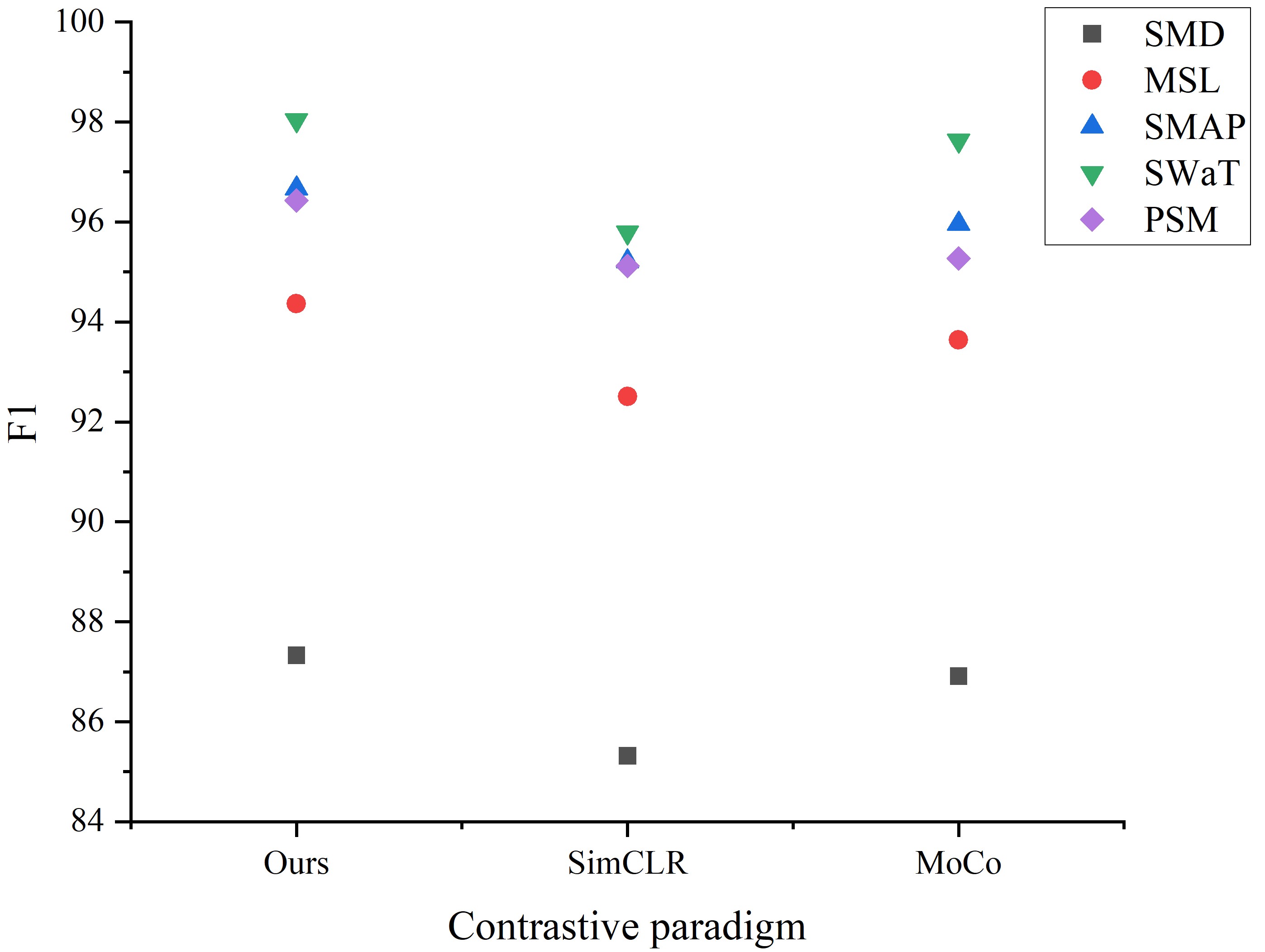} \label{contrastiveF1}}
\subfloat[The training time for each epoch]{
		\includegraphics[scale=0.26]{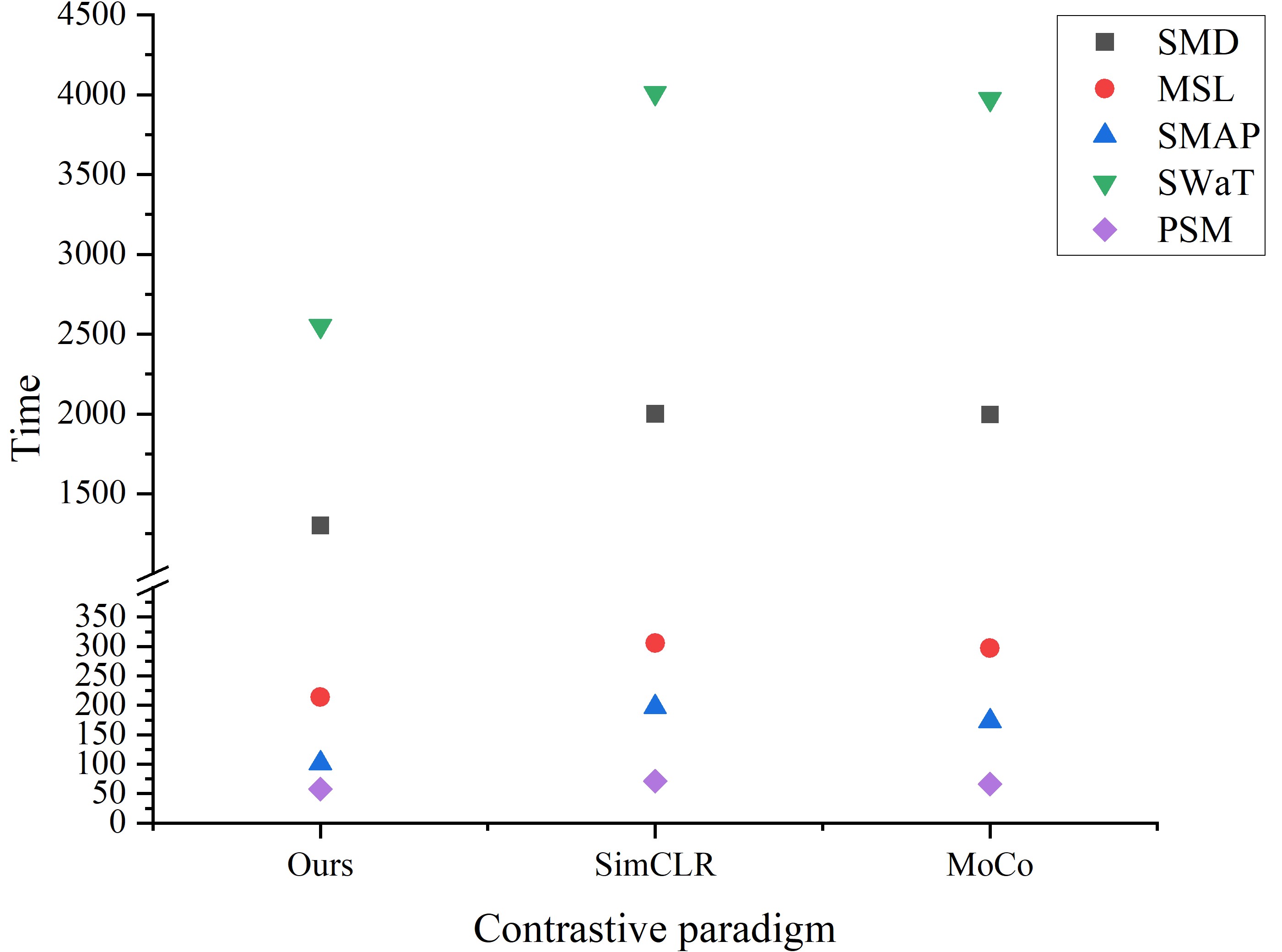} \label{contrastiveTime}}
\caption{The ablation experiments of contrastive paradigms}
\label{fig: contrastive ablation}
\end{figure*}

In addition, to further demonstrate the effectiveness of our proposed contrastive learning approach, we conducted ablation experiments using two widely used contrastive learning methods:  SimCLR \citep{chen2020simple} and MoCo \citep{he2020momentum}. Unlike these methods, which generate negative samples, TransDe relies solely on positive samples to compute the loss and optimize parameters. In our implementation of SimCLR and MoCo, we treated two patch-level representations of the same sample as positive pairs, while considering representations from different samples within the same batch as negative pairs. The results are presented in Fig. \ref{fig: contrastive ablation}. As shown in Fig. \ref{contrastiveF1}, our proposed method consistently outperforms both SimCLR and MoCo across all datasets, demonstrating its effectiveness and superiority. Furthermore, Fig. \ref{contrastiveTime} illustrates that our approach has the lowest time complexity. This advantage arises from the fact that MoCo and SimCLR require the generation of negative sample pairs, which involves more computationally intensive operations.

\subsection{Model analysis}

In this section, we analyze how the choice of patch size affects model performance. Different patch sizes provide multi-scale attention information, which can enhance the model's ability to capture important features. We take SMAP, SWaT, and PSM datasets as examples, with results presented in Table \ref{patchsize}. To prevent information leakage, we select odd numbers for the patch sizes. Our observations indicate that using multi-scale patch sizes improves the model's performance by enabling the extraction of more valuable information for feature representation. It is important to note that the optimal patch size varies by dataset. For SMAP, the best selection of patch size is $[3, 5]$, while for SWaT and PSM datasets, the best choice is $[1, 3, 5]$.
\begin{table*}[!htbp]
	\centering
	\caption{The analysis experiments of TransDe on patch level. The best results are in bold. All results are in \%.}
	\label{patchsize}
 \resizebox{\textwidth}{!}{
	\begin{tabular}{c | c c c | c c c | c c c }
		\hline
		Datasets & \multicolumn{3}{c|}{SMAP} & \multicolumn{3}{c|}{SWaT} & \multicolumn{3}{c}{PSM}\\\hline
            Patch size & P & R & F1 & P & R & F1 & P & R & F1\\\hline
            $\lbrack$1$\rbrack$ & 95.46 & 94.26 & 94.86 & 93.21 & 94.28 & 93.74 & 92.85 & 94.74 & 93.76\\
            $\lbrack$3$\rbrack$ & 96.01 & 94.75 & 95.38 & 95.11 & 94.96 & 95.03 & 91.55 & 96.37 & 93.90\\
            $\lbrack$5$\rbrack$ & 94.37 & 95.82 & 95.09 & 95.39 & 96.01 & 95.70 & 95.84 & 92.41 & 94.09\\
            $\lbrack$1, 3$\rbrack$ & 95.21 & 97.85 & 96.51 & 96.53 & 93.77 & 95.13 & 96.42 & 95.11 & 95.76\\
            $\lbrack$1, 5$\rbrack$ & 96.99 & 95.41 & 96.19 & 92.74 & 93.85 & 93.29 & 93.79 & 94.88 & 94.33\\
            $\lbrack$3, 5$\rbrack$ & 96.83 & 97.16 & $\mathbf{96.99}$ & 92.75 & 96.85 & 94.76 & 96.14 & 96.23 & 96.18\\
            $\lbrack$1, 3, 5$\rbrack$ & 94.35 & 97.68 & 95.98 & 95.86 & 96.41 & $\mathbf{96.13}$ & 93.13 & 98.46 & $\mathbf{96.43}$\\\hline
	\end{tabular}}
\end{table*}

\begin{figure*}[!htb]
\centering
\subfloat[The F1 metric]{
		\includegraphics[scale=0.25]{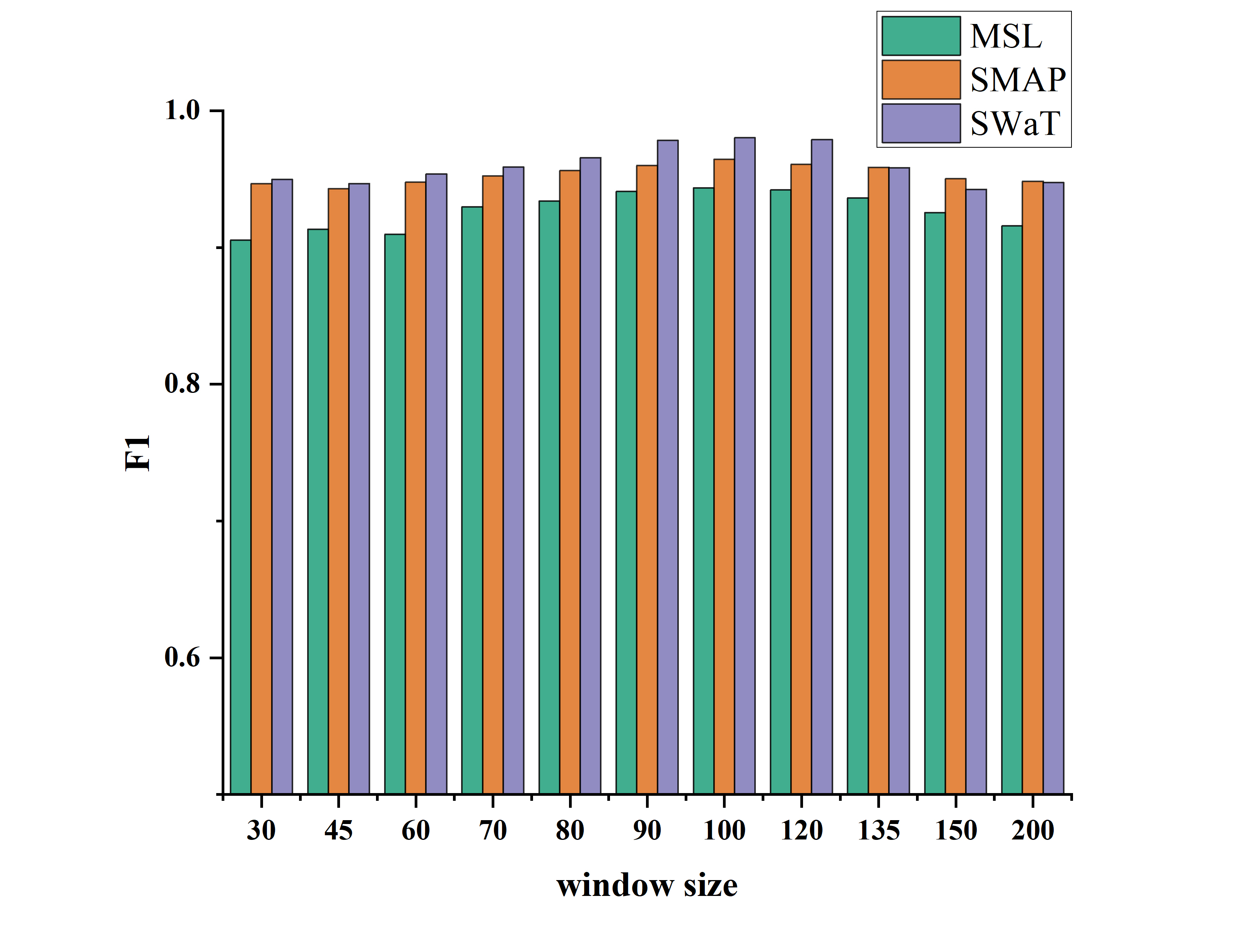} \label{f1}}
\subfloat[The ACC metric]{
		\includegraphics[scale=0.25]{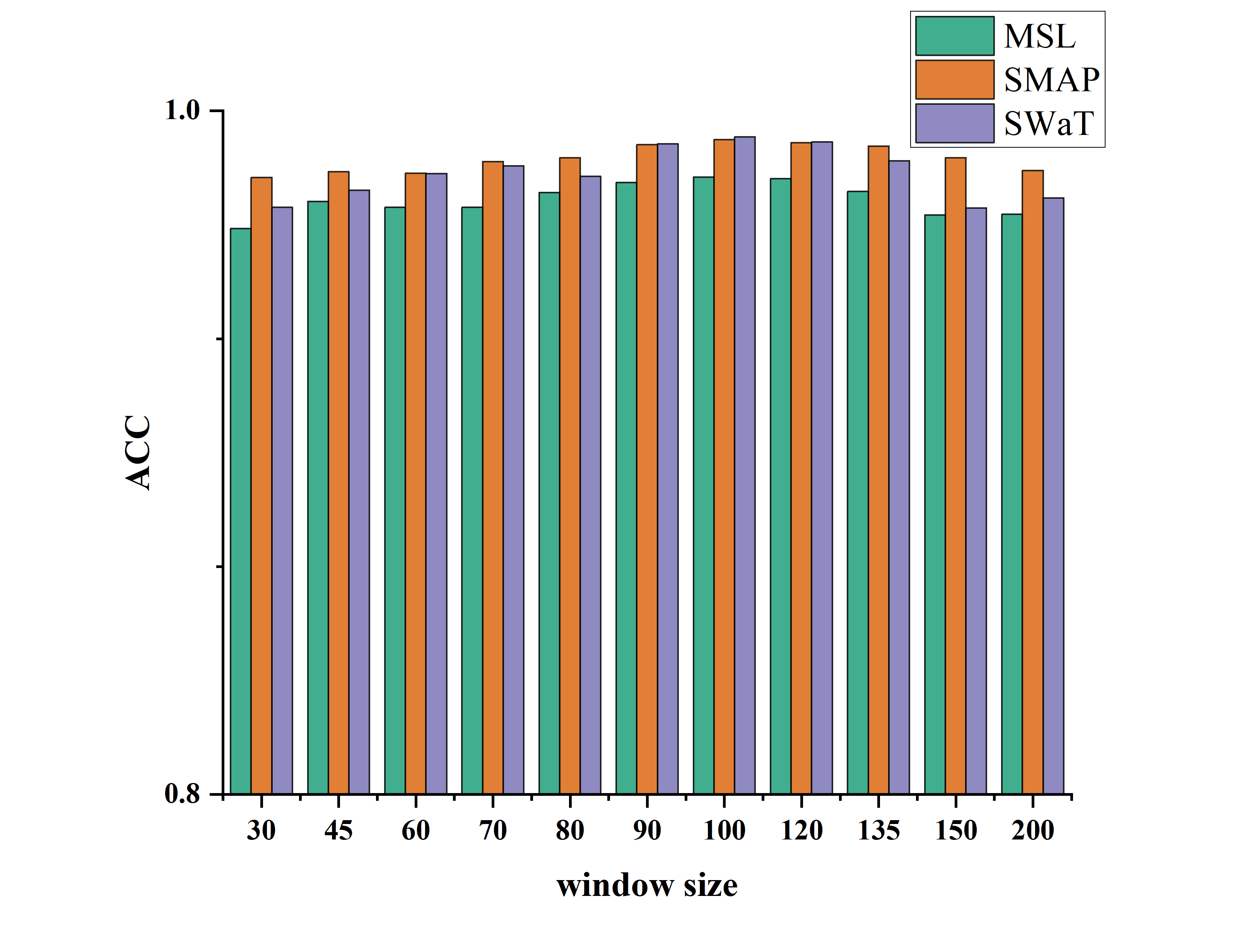} \label{acc}}
\caption{The model performance of experiments on window size}
\label{window size}
\end{figure*}

Second, we examine how window size affects the model's performance. Window size is crucial for time series segmentation, providing historical context for the target timestamp. The results are shown in Fig. \ref{window size}. From the observations, it can be concluded that window size significantly impacts TransDe's performance. Specifically, using petite window sizes (such as 30 or 45) or excessively large ones (like 150 or 200) can lead to performance degradation. Small window sizes may fail to capture essential dependency information, while large window sizes often include excessive redundant data and noise, potentially causing overfitting. Additionally, larger window sizes increase memory usage and computational complexity. Therefore, selecting an appropriate window size helps avoid information overload while maintaining computational efficiency.

\subsection{Hyperparameter sensitivity}
\begin{figure*}[!htb]
\centering
\subfloat[Layers]{
		\includegraphics[scale=0.2]{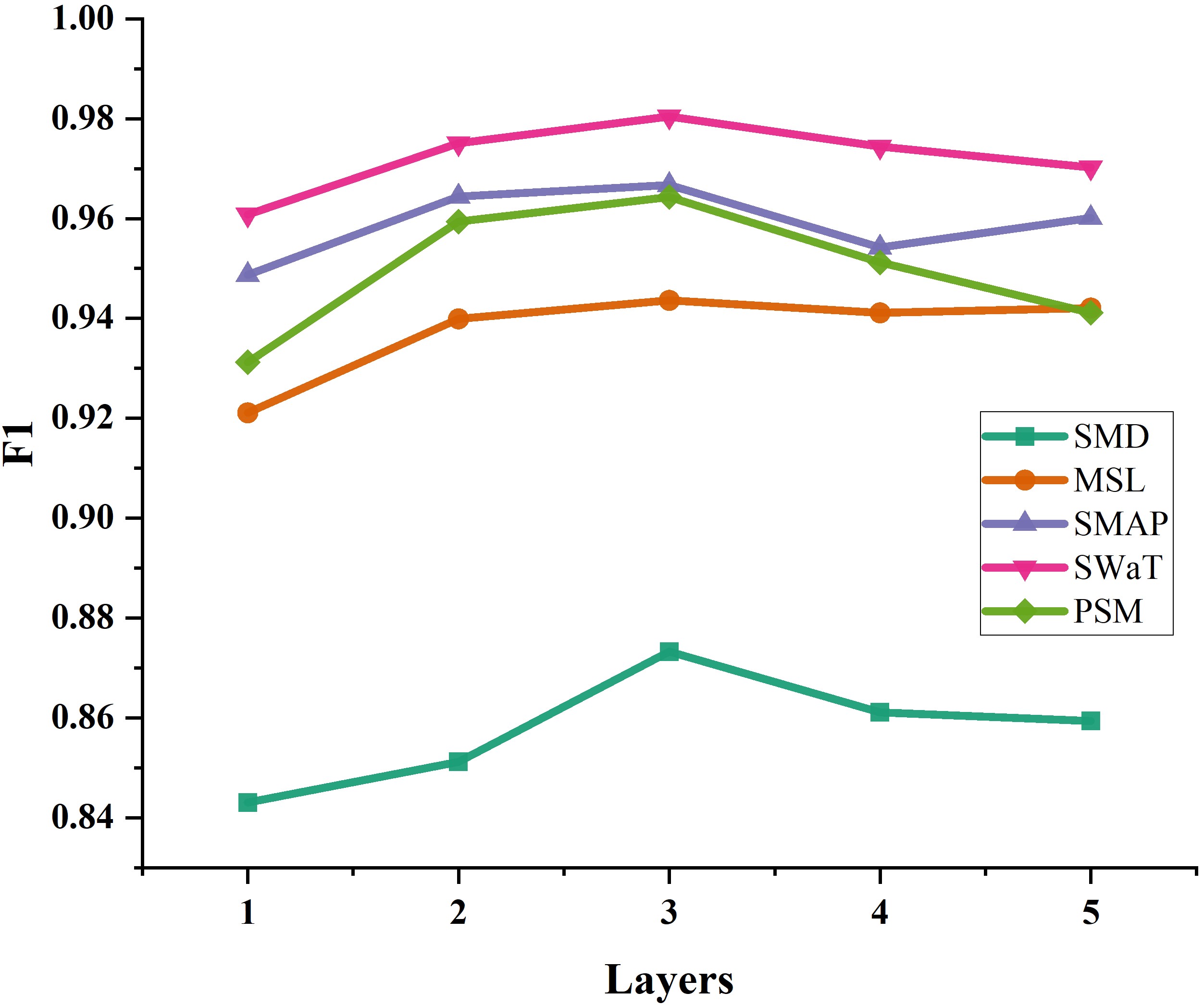} \label{layers}}
\subfloat[Hidden size]{
		\includegraphics[scale=0.2]{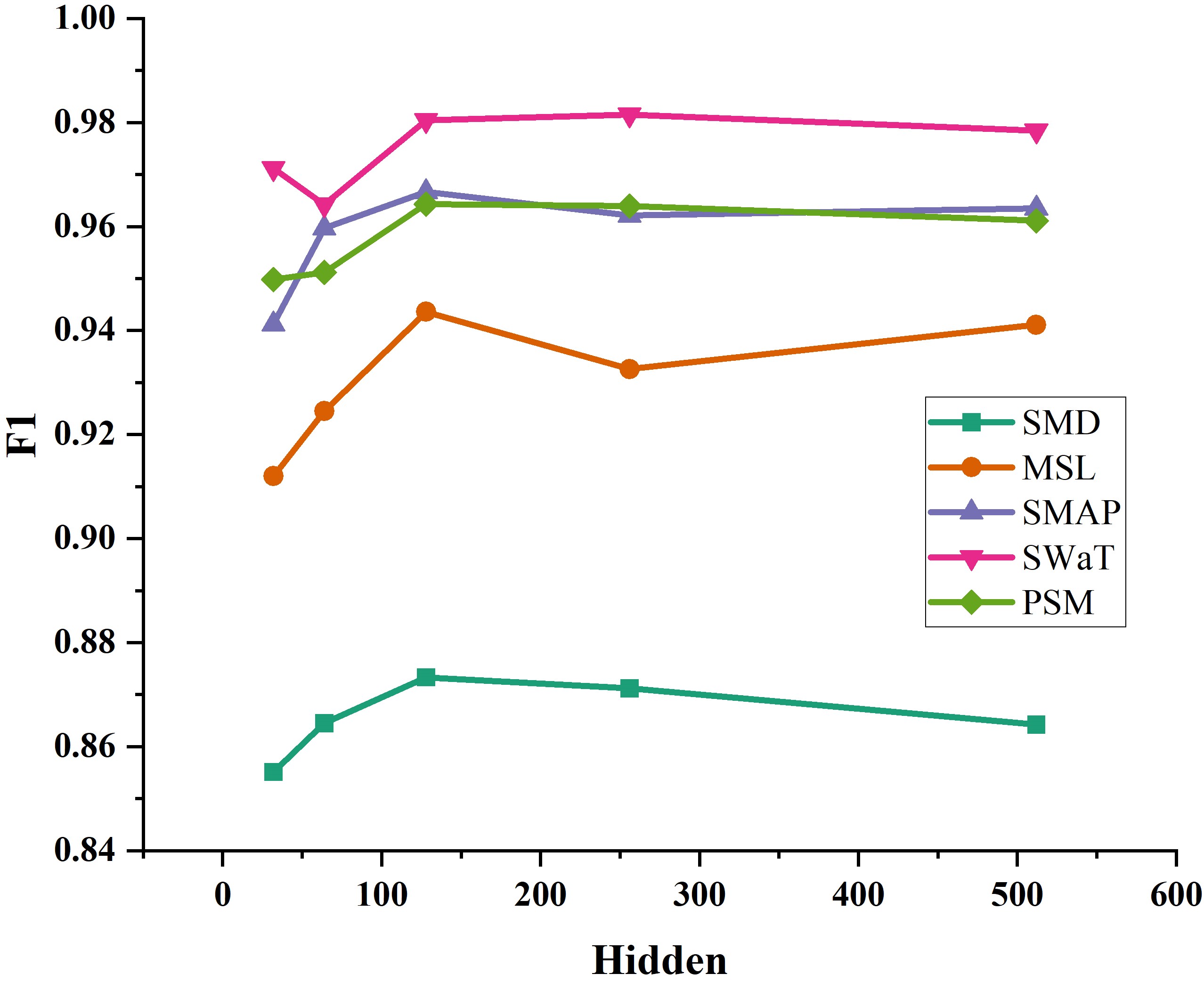} \label{hidden}}
  \subfloat[Heads number]{
		\includegraphics[scale=0.2]{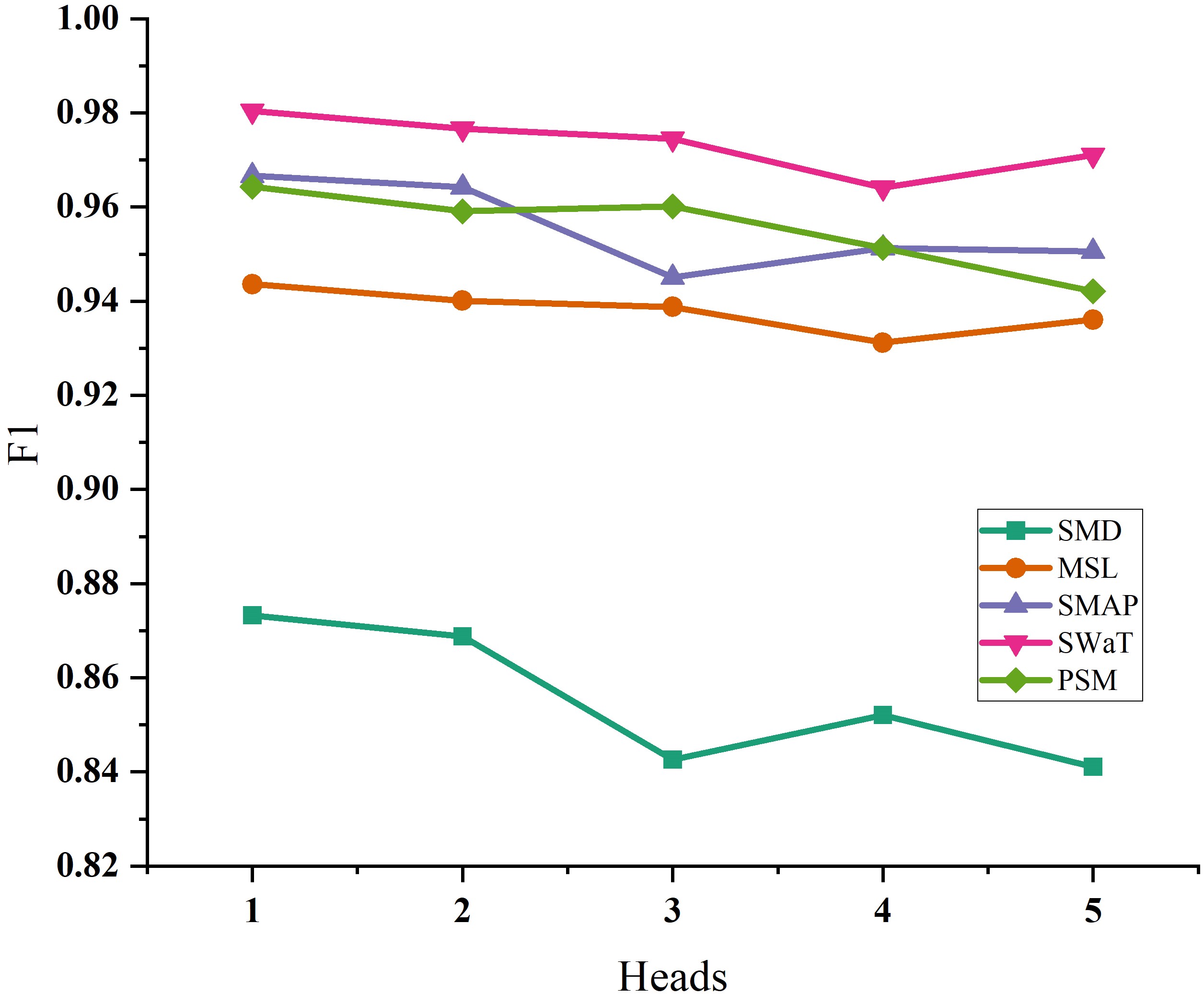} \label{heads}}
\caption{The observations of hyperparameter sensitivity experiments on five datasets}
\label{Hyperparameter sensitivity}
\end{figure*}

In this section, we conduct experiments on the setting of hyperparameters, including the layers of the transformer, the hidden size of hidden size in the model, and the number of heads in the attention mechanism, as shown in Fig. \ref{Hyperparameter sensitivity}. 

Specifically, we vary the number of transformer layers from 1 to 5, with the results displayed in Fig. \ref{layers}. Our findings indicate that using too many layers can lead to a slight decrease in performance. This decline is likely due to excessive processing, which can obscure the representative features of the original sequences. Conversely, using only a single layer may not provide sufficient capacity to extract meaningful information. Based on these observations, we conclude that the optimal number of transformer layers is 3.

We also explore the impact of hidden size by testing values of 64, 128, 256, and 512. The results, shown in Fig. \ref{hidden}, reveal that both excessively large and small hidden sizes do not improve performance. A large number of hidden units can strain the computational process and waste resources, while too few units diminish TransDe's ability to capture relevant information. Thus, we determine that a hidden size of 128 is most effective after thorough analysis.

To investigate the effect of the number of heads in the attention mechanism, we test values ranging from 1 to 5, as illustrated in Fig. \ref{heads}. Our observations show that as the number of heads increases, model performance tends to degrade. This decline may result from multi-head attention causing the model to lose focus on the global representation, introducing more noise into the learning process. Consequently, we set the number of heads in the attention mechanism to 1.

\subsection{Visualization}

\begin{figure*}[!htb]
    \centering
    \includegraphics[width=12cm,height=8cm]{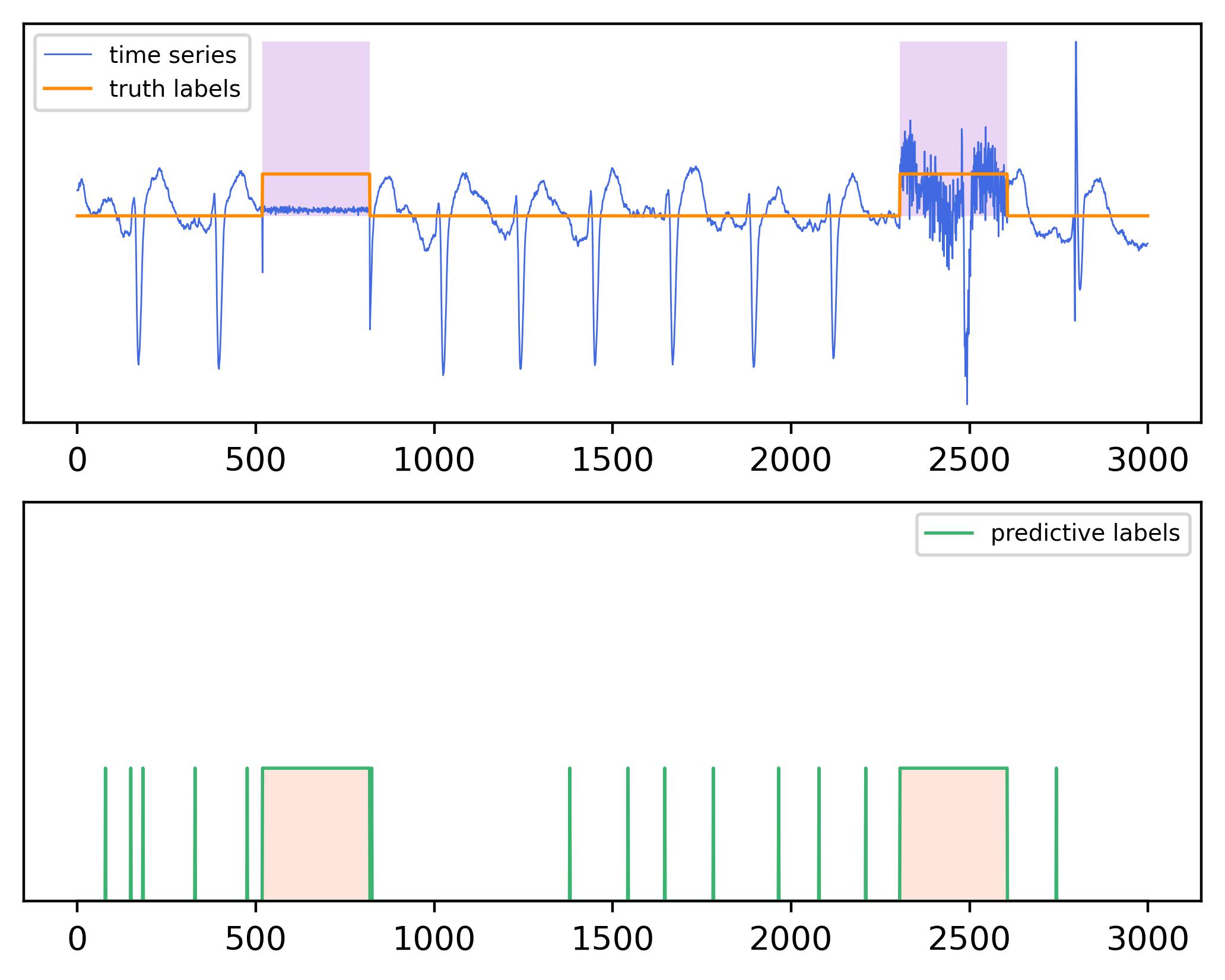}
    \caption{Visualization}
    \label{fig: Visualization}
\end{figure*}

To intuitively evaluate the validity of TransDe in identifying anomalies, we visualize the original sequence alongside the predicted scores. Specifically, we focus on a subset of the UCR univariate time series \citep{DCdetector}, examining timestamps from 2000 to 5000. The results of this visualization are presented in Fig. \ref{fig: Visualization}.

The figure shows that anomalies are present in the original data, specifically between timestamps 500 to 800 and 2300 to 2600, and these are successfully identified in the predicted labels. This highlights the effectiveness of TransDe in detecting anomalies. However, there are instances of incorrect predictions, particularly in the normal samples from timestamps 0 to 500, where some are mistakenly classified as anomalies. In summary, while there is room for improvement, TransDe demonstrates a relatively strong ability to identify anomalies.

\section{Conclusion}
\label{Conclusion}
This paper proposes a novel transformer framework based on decomposition (TransDe) for multivariate time series anomaly detection. This approach addresses the limitations of reconstruction-based methods, which are often affected by noise in the data. TransDe begins by employing a decomposition operation to extract various components from the original time series, which can extract more typical sequence characteristics for representation learning. Then it leverages patch operation to divide each sequence component into patches and models both inter-patch and intra-patch dependency representations using a transformer architecture. The framework then integrates these representations according to the patch level and employs a pure contrastive loss function to learn the latent dependencies within the time series. Extensive experiments demonstrate that TransDe outperforms twelve baseline methods across five public datasets.

For future work, further experiments will be executed with diverse datasets from various industries to validate the broader applicability of TransDe. In addition, given the potential of the frequency domain in enhancing anomaly detection for time series data, we aim to explore more efficient and innovative methods that leverage both time and frequency information to improve performance.
\bibliographystyle{apalike} 
\bibliography{ref.bib}



\end{document}